\documentclass[journal ]{new-aiaa}
\usepackage[utf8]{inputenc}

\usepackage{textcomp}

\usepackage{graphicx}
\usepackage{amsmath}
\usepackage{lipsum}
\usepackage[version=4]{mhchem}
\usepackage{siunitx}
\usepackage{longtable,tabularx}
\usepackage{caption}
\usepackage{subcaption}
\usepackage{float}
\usepackage{enumitem}
\usepackage{xcolor}
\usepackage{gensymb}
\usepackage{multirow}
\usepackage{hyperref}
\usepackage{color, colortbl}
\usepackage{algpseudocode}
\usepackage{algorithm}
\definecolor{Gray}{gray}{0.9}
\definecolor{LightCyan}{rgb}{0.88,1,1}
\usepackage[english]{babel}

\usepackage{amsthm}
\usepackage{mathtools}

\usepackage{amssymb}
\usepackage{optidef}

\usepackage{mathabx}
\newtheorem{theorem}{Theorem}[section]
\newtheorem{lemma}[theorem]{Lemma}

\title{Energy Optimal Traversal Between Hover Waypoints for Lift+Cruise Electric Powered Aircraft}
\date{June 2024}
\author{Akshay Mathur\footnote{PhD Candidate, Robotics Institute, University of Michigan, AIAA Student Member}}
\affil{University of Michigan, Ann Arbor, MI, 48109}

\author{ Ella Atkins \footnote{Fred D. Durham Professor and Head, Kevin T. Crofton Aerospace and Ocean Engineering Department, Virginia Tech, AIAA Fellow} }
\affil{Virginia Tech, Blacksburg, VA 24061}

\begin{document}

\maketitle

\begin{abstract}

Advanced Air Mobility aircraft require energy efficient flight plans to be economically viable. This paper defines minimum energy direct trajectories between waypoints for Lift+Cruise electric Vertical Take-Off and Landing (eVTOL) aircraft. 
Energy consumption is optimized over accelerated and cruise flight profiles with consideration of mode transitions. 
Because eVTOL operations start and end in hover for vertical take-off and landing, hover waypoints are utilized. 
Energy consumption is modeled as a function of airspeed for each flight mode, providing the basis to prove energy optimality for multi-mode traversal.  
Wind magnitude and direction dictate feasibility of straight-line traversal because Lift+Cruise aircraft point into the relative wind direction while hovering but also have a maximum heading rate constraint.
Energy and power use for an experimentally validated QuadPlane small eVTOL aircraft are characterized with respect to airspeed and acceleration in all flight modes. Optimal QuadPlane traversals are presented.
Constraints on acceleration and wind are derived for straight-line QuadPlane traversal.
Results show an optimal QuadPlane $500m$ traversal between hover waypoints saves $71\%$ energy compared to pure vertical flight traversal for a representative case study with a direct $4m/s$ crosswind. 
Energy optimal eVTOL direct trajectory definition with transitions to and from hover is novel to this work.
Future work should model three-dimensional flight and wind as well as optimize maneuver primitives when required.

\end{abstract}

\section*{Nomenclature}
{\renewcommand\arraystretch{1.0}
\noindent\begin{longtable*}{@{}l @{\quad=\quad} l@{}}
$\Vec{V}^g, V^g$ & Ground frame velocity vector and magnitude (ground speed)\\
$\Vec{V}^a, V^a$ & Body frame velocity vector and airspeed\\
$u,v,w$ & Body frame velocity components \\
$V_x, V_y, V_z$ & Ground frame velocity components\\
$x_B, y_B, z_B$ & Axes for Body frame\\
$x_E, y_E, z_E$ & Axes for Earth / Ground frame\\
$x_A, y_A, z_A$ & Axes for Air / Wind frame\\
$\alpha, \beta, \gamma$ & Angle of attack, Sideslip angle and Flight path angle for the aircraft, respectively\\
$\sigma, \chi$ & Aircraft heading angle and Course angle, respectively\\
$V^w, \sigma^w$ & Lateral wind velocity magnitude and heading angle\\
$\prescript{B}{}{\textbf{R}_E}$, $\prescript{E}{}{\textbf{R}_B}$ & Earth to Body frame and Body to Earth frame rotation matrices \\ 
$Q,P,H$ & $Quad$ (vertical), $Plane$ (cruise) and $Hybrid$ (transition) flight modes for an eVTOL aircraft\\
$\Vec{w}_i$ & Hover waypoint\\
$x_i, y_i, z_i$ & Ground frame coordinates of $\Vec{w}_i$\\
$l_{i,i+1}$ & Distance between hover waypoints $\Vec{w}_i$ and $\Vec{w}_{i+1}$\\
$\mathbb{T}_{i,i+1}, \mathbb{T}_{i,i+1}^w$ & General straight and level traversal in the absence and presence of steady wind\\
$k$ & Indicates acceleration "$-$" or deceleration "$-$"\\
$\mathcal{T}_k^w, \mathcal{T}_k$ & Cubic spline ground speed profile for accelerated flight with and without steady wind\\
$V^a_c, V^g_c$ & Cruise airspeed and ground speed for the aircraft\\
$a^g_{max,k}$ & Maximum acceleration over the cubic spline ground speed profile\\
$a^a_{lim,k}$ & Aircraft's acceleration/deceleration performance limit\\
$V_{QH}$ & Switching airspeed between "Quad" (vertical) and "Hybrid" (transition) flight modes\\
$V_{HP}$ & Switching airspeed between "Hybrid" (transition) and "Plane" (forward/cruise) flight modes\\
$t_k, l_k$ & Time duration and Distance traveled over the accelerated segment $\mathcal{T}_k$, respectively\\
$l_c$ & Distance traveled over the cruise segment\\
$V_{max}$ & Maximum achievable airspeed over the traversal $\mathbb{T}_{i,i+i}$\\
$l_{min}^V$ & Minimum distance between hover waypoints to reach airspeed $V$ over the traversal $\mathbb{T}_{i,i+1}$\\
$E_{\mathbb{T}}$ & Energy consumed over the traversal $\mathbb{T}_{i,i+1}$\\
$E(\mathcal{T}_k), E(V_c)$ & Energy consumed over the accelerated segment $\mathcal{T}_k$ and over the cruise segment\\
$\mathcal{F}(V_c)$ & Energy consumed per distance traveled while cruising at $V_c$ in no wind\\
$P(V_c)$ & Power required to maintain cruise speed $V_c$\\
$V^*_c$ & Most energy efficient cruise speed in no wind\\
$V_{lim}$ & Maximum achievable cruise speed in no wind\\
$V_{stall}$ & Aerodynamic stall speed for the aircraft in "Plane" Mode\\
$B^{cr}, S^{cr}$ & Set of boundary critical points and solution critical points, respectively\\
$\Dot{\sigma}_{lim}$ & Aircraft's heading rate performance limit\\
$\Delta\sigma$ & Relative wind direction, i.e. $\Delta \sigma = \chi - \sigma^w$\\
$E_\mathbb{T}^w$ & Energy consumed over the traversal $\mathbb{T}_{i,i+q}^w$ in steady wind\\
$E(\mathcal{T}_k^w), E(V^g_c)$ & Energy consumed over the accelerated segment $\mathcal{T}_k^w$; and cruise segment in steady wind\\
$STF$ & Straight-line Traversal Feasibility between hover waypoints in steady wind\\
$\prescript{}{min}{a^g_{max,k}}$ & Minimum value set for the maximum acceleration over cubic spline ground speed profile $\mathcal{T}_k^w$\\
$\prescript{}{min}{V^g_c}$ & Minimum value set for the cruise ground speed for traversal $\mathbb{T}_{i,i+1}^w$\\
$t^w_k, l_k^w$ & Time duration and distance traveled over the accelerated segment $\mathcal{T}^w_k$ in steady wind\\
${V^a_c}^*, {V^g_c}^*$ & Most energy efficient cruise airspeed and ground speed, respectively

\end{longtable*}}

\section{Introduction}

Advanced Air Mobility (AAM) \cite{AAM_NASA, AAM_taxonomy} promises fast and convenient transportation of goods and people with increasingly autonomous \cite{paths_to_uam} electric vertical take-off and landing (eVTOL) vehicles. AAM missions include on-demand air mobility \cite{uber_UAM_whitepaper, AAM_missions} for passengers and heavy cargo as well as small Uncrewed Aircraft System (sUAS) operations such as mapping \cite{multicopter_landslide_mapping}, surveillance \cite{traffic_monitoring_with_multicopters}, inspection \cite{drone_inspection_bridge}, and package delivery \cite{drone_package_delivery}.
Several AAM vehicle designs have been proposed \cite{VFS_eVTOL_directory} and are broadly categorized \cite{uam_vtol_types, uam_VTOL_types_2,UAM_vehicles_baseline_2019, tilt_wing_concept} as Multirotor, Side-by-Side Helicopter, Lift+Cruise and Tilt-wing. Lift+Cruise eVTOL offers energy efficient cruise plus short-term hover capability.
AAM eVTOL aircraft have three modes of operation - vertical (lift), forward (cruise) and transition flight. 
An AAM flight plan requires taking off vertically, then transitioning to forward flight mode while continuing acceleration to cruise speed. Near the destination, the eVTOL aircraft decelerates back to vertical flight mode to prepare for a vertiport landing. eVTOL aircraft, e.g., sUAS, may need to hover at specific waypoints for missions such as surveillance and inspection, resulting in multiple trajectory segments that require transitions from vertical to forward flight mode and vice versa.


Early vertical take-off and landing (VTOL) aircraft such as the V-22 Osprey \cite{V22_aero_development} required the flight crew to manually trigger flight mode transitions, but the transition sequence itself was fully automated to avoid aerodynamic instabilities inherent in the twin engine V-22 tilt-rotor design \cite{V22_transition_challenges}. 
VTOL aircraft aerodynamic modeling and control design has since advanced by combining fixed-wing aircraft \cite{stevens_aircraft_2015, beard_and_mclain} and multicopters \cite{quan_quan} enabling autonomous transition control schemes as shown in \cite{QP_conference_paper, ddp_mpic_lift+cruise}. 
Autonomous transitions are necessary for ensuring safe and reliable operations but require accurate aircraft models for designing robust transition control schemes.
Aerodynamic data is essential for developing aircraft models, however despite the release of several designs \cite{VFS_eVTOL_directory}, experimentally validated aerodynamic data for eVTOL aircraft is not easily accessible with our small Uncrewed Aircraft System (sUAS) QuadPlane as an exception \cite{QP_arXiv_journal_paper, QP_JOA_engineering_note}. 
Trajectory planning for fixed wing and multicopter sUAS has been done in zero wind \cite{traj_planning_aircraft_no_wind, traj_planning_multicopter_no_wind}, yet some sUAS have operating airspeeds comparable to typical wind speeds. 
Energy-efficient trajectory generation in a wind field for a fixed-wing sUAS \cite{spline_energy_eff_traj,traj_plan_mav_in_wind} as well as for multicopters \cite{quad_flight_in_wind} has also been studied, but not for a transitioning eVTOL. 
Authors in \cite{stop-n-go_quadcopter} study energy efficiency of stop-n-go trajectories between hover waypoints using a quadcopter. 
Because AAM aircraft hover consumes far more energy than wing-lift-supported cruise flight, Lift+Cruise eVTOL aircraft will typically transition to cruise flight whenever possible.  
Optimal control approaches are applied in \cite{ddp_mpic_lift+cruise} using NASA's Lift+Cruise aircraft concept \cite{uam_VTOL_types_2}, yet pure cruise mode is not utilized as vertical motors are spinning even at cruise airspeeds.
This paper examines how Lift-Cruise eVTOL energy savings can be further improved with optimal flight mode switching between lift, cruise and transition flight modes.

This paper presents a novel analysis of accelerated flight mode switching to derive energy optimal traversal between hover waypoints for Lift+Cruise aircraft in steady wind. 
To derive an energy optimal traversal solution, we first formally define all vehicle specific parameters affecting flight mode selection and straight-line acceleration, cruise and deceleration profiles. 
Axioms characterize energy consumption as a function of airspeed for Lift+Cruise aircraft across all three flight modes consistent with experimental data in \cite{QP_arXiv_journal_paper}.
Energy optimality is proven in zero wind for our multi-mode traversal solution over the set of feasible acceleration, deceleration and cruise airspeed values. 
Straight-line traversal is desired as the minimum distance solution, but in wind Lift+Cruise aircraft hover stably by pointing into the relative wind. 
Because relative wind direction rapidly changes with airspeed over accelerated flight, the feasibility of straight-line traversal depends on wind magnitude, direction, aircraft acceleration and heading rate limit. 
When straight-line traversal is infeasible, the use of maneuver primitives \cite{maneuver_primitive} is prescribed. An example cubic spline maneuver primitive is described over course angle around each hover waypoint to assure satisfaction of the aircraft heading rate constraint. An eVTOL sUAS QuadPlane \cite{QP_conference_paper} platform case study is explored due to its publicly available aerodynamic data \cite{QP_arXiv_journal_paper, QP_JOA_engineering_note}.  
Note that QuadPlane \cite{QP_conference_paper} mode nomenclature is adopted in this work, i.e. vertical (lift) is "$Quad$" mode, forward (cruise) is "$Plane$" mode, and transitions occur in "$Hybrid$" mode.


The primary contribution of this paper is definition of energy optimal traversal between adjacent hover waypoints for Lift+Cruise aircraft. Previous work has derived energy optimal trajectories with constant speed \cite{traj_plan_mav_in_wind, quad_flight_in_wind} and accelerated flight \cite{spline_energy_eff_traj, ddp_mpic_lift+cruise} but assumes operations in only a single flight mode.
Our second contribution is definition of a feasibility constraint for straight-line traversal in a given steady wind magnitude and direction subject to aircraft acceleration and heading rate constraints. 
The final contribution of this paper is specification of power, energy, and energy optimal traversal profiles for experimentally validated \cite{QP_arXiv_journal_paper} QuadPlane sUAS \cite{QP_conference_paper}. Explicit constraints on wind magnitude and direction partition cases for straight-line traversal versus maneuver primitive traversal in steady wind.

Below, Section \ref{Section: Prelims} defines coordinate frames and relevant aircraft state and wind parameters.  Section \ref{Section: Optimal Traversal in No wind} defines a straight-line traversal data structure and its specification to minimize energy cost. Next, Section \ref{Section: Optimal traversal in steady wind} describes traversal in steady wind, including an algorithm to determine straight-line traversal feasibility and an example maneuver primitive to apply when necessary. Section \ref{Section: QP case study} describes energy optimal traversal for the QuadPlane use case followed by concluding remarks in Section \ref{Section: Conclusion}.

\section{Preliminaries}
\label{Section: Prelims}

Aircraft motion occurs in three dimensions with three translational and three rotational degrees of freedom. The aircraft translational velocity vector over the ground ($\Vec{V}=\Vec{V^g}$) can be expressed in either an Earth-fixed ($E$) or Body-fixed ($B$) frame. Frame $E$ adopts a North - East - Down ($NED$) convention, and frame $B$ uses a Forward - Right - Down ($FRD$) convention as in \cite{stevens_aircraft_2015,McClamroch}. Body frame coordinates can be derived from the Earth frame by executing a yaw ($\psi$) - pitch ($\theta$) - roll ($\phi$) 3-2-1 Euler angle sequence. Wind or air frame $A$ has the same origin as $B$ but is oriented with $x_A$ pointing into the relative wind.   
Body frame velocity is defined as $\prescript{B}{}{\Vec{V}^a} = \begin{bmatrix}
    u & v & w
\end{bmatrix}^T$.  This paper assumes $u$, $v$, $w$ are estimated from the aircraft air-data system and denoted with superscript $a$. 
The inertial velocity vector in the Earth frame is defined as $\prescript{E}{}{\Vec{V}^g} = \begin{bmatrix}
    V_x & V_y & V_z
\end{bmatrix}^T$. Magnitudes of relative wind and ground velocity are defined by $V^a = ||\prescript{B}{}{\Vec{V}^a} ||_2 =\sqrt{u^2 + v^2 + w^2}$ and $V^g = ||\prescript{E}{}{\Vec{V}^g}||_2= \sqrt{V_x^2 + V_y^2 + V_z^2}$, respectively. In zero ambient wind, airspeed and ground speed are equal, i.e. $V^a = V^g$. 

\begin{figure}[!ht]
    \centering
    \begin{tabular}{cc}
        \includegraphics[width = 0.55\linewidth]{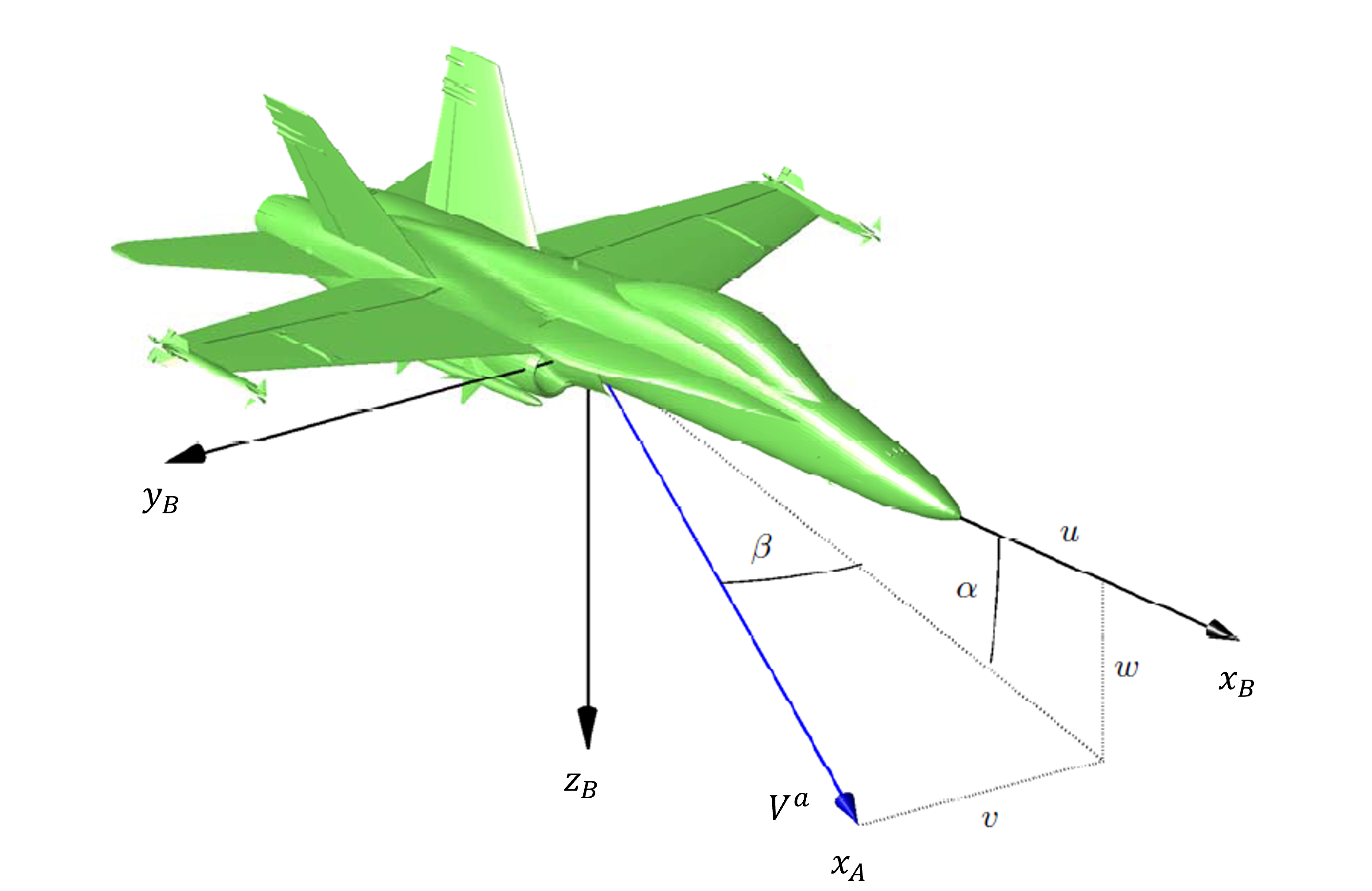} & 
        \includegraphics[width = 0.39\linewidth]{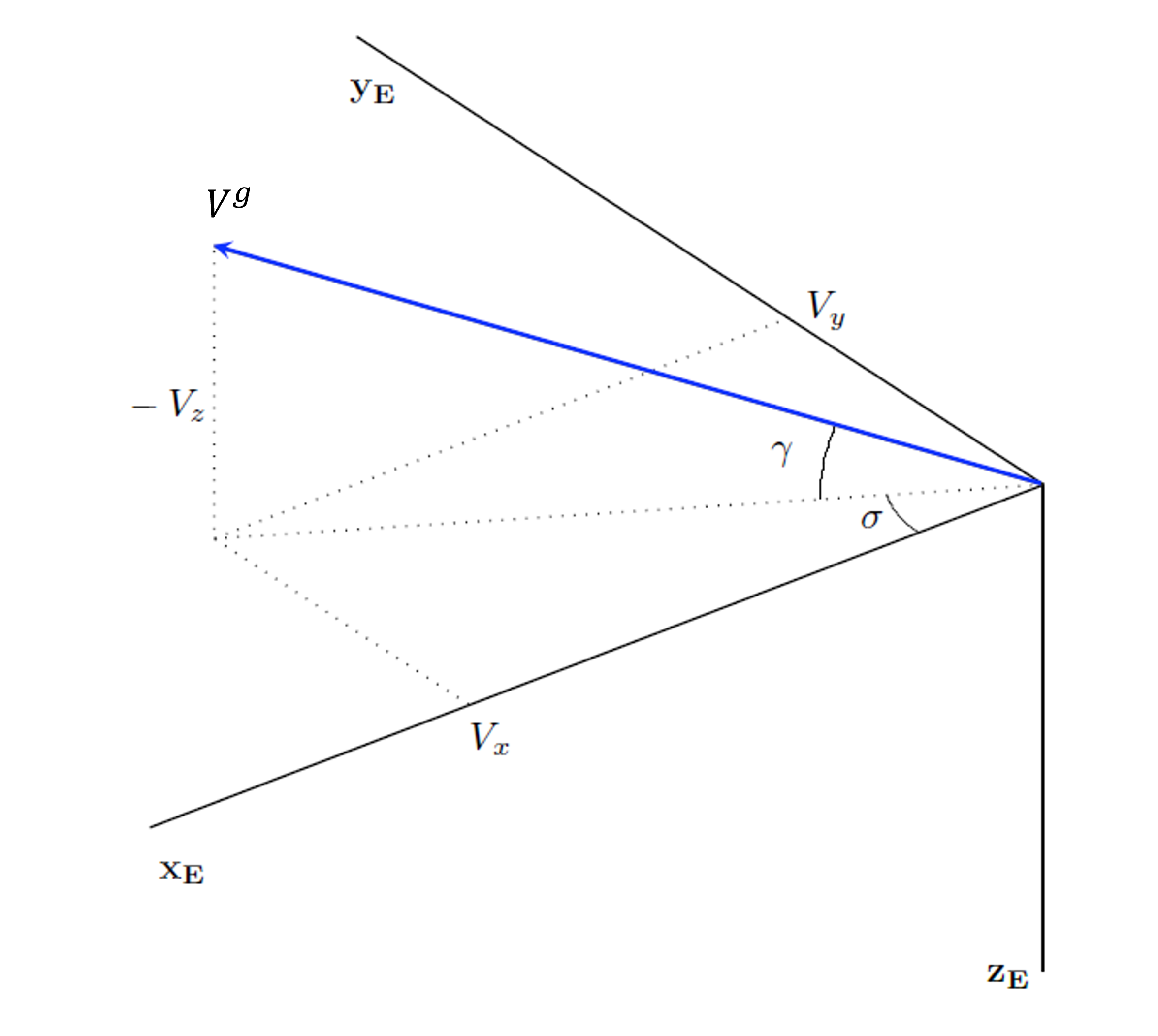}
    \end{tabular}
    \caption{Aircraft velocity in Body (left) and Earth (right) coordinates \cite{McClamroch}. Relative wind is also shown (left).}
    \label{fig: Aircraft velocity in Body Frame and Earth Frame.}
\end{figure}

Flight path $\gamma$ is the angle between the ground velocity vector and Earth frame horizontal plane. Aircraft heading $\sigma$ is the angle between the projection of the ground velocity vector onto the horizontal plane and the $x_E$ axis. Components of the ground velocity vector are:

\begin{equation}
    \begin{aligned}
        V_x &= V^g cos(\gamma) cos(\sigma); &&
        V_y &= V^g cos(\gamma) sin(\sigma); &&
        V_z &= -V^g sin(\gamma).
    \end{aligned}
    \label{eq: aircraft velocity in ground frame}
\end{equation}

The wind vector relative to body frame $B$ is defined by airspeed $V^a$, angle of attack $\alpha$, and sideslip angle $\beta$. This relative wind vector defines aerodynamic forces and moments. Components of the body frame velocity relative to the wind are given as:

\begin{equation}
    \begin{aligned}
        u &= V^a cos(\alpha) cos(\beta); && 
        v &= V^a cos(\alpha) sin(\beta); &&
        w &= V^a sin(\alpha) 
    \end{aligned}
\end{equation}

\noindent Angles $\alpha$ and $\beta$ can be determined from $\prescript{B}{}{\Vec{V}^a}$ as follows:
\begin{equation}
\begin{aligned}
    tan (\alpha) &= \frac{w}{\sqrt{u^2 + v^2}}; &&
    sin(\beta) &= \frac{v}{\sqrt{u^2 + v^2}}
\end{aligned}
\end{equation}
\noindent Flight path angle and aircraft heading can be determined from ground velocity as follows: 
\begin{equation}
\begin{aligned}
    tan (\gamma) & = \frac{-V_z}{\sqrt{V_x^2 + V_y^2}}; && 
    sin (\sigma) & = \frac{V_y}{\sqrt{V_x^2 + V_y^2}}
\end{aligned}
\end{equation}

Consider aircraft motion in steady horizontal wind. Define wind magnitude as ${V}^w$ and wind heading as $\sigma^w$. By convention, wind blowing from South to North has $\sigma^w = 0$.
The direction of horizontal aircraft ground velocity in the Earth frame is defined as course angle $\chi$, i.e. $tan(\chi) = \left( \frac{V_y}{V_x} \right)$.
Wind influences airspeed such that relative wind vector ($\prescript{B}{}{\Vec{V}^a}$) is given by:

\begin{equation}
    \begin{aligned}
        \prescript{B}{}{\Vec{V}^a} &= \begin{bmatrix}
            u \\
            v \\ 
            w 
        \end{bmatrix}& = \begin{bmatrix}
            V cos(\alpha) cos(\beta)\\
            V cos(\alpha) sin(\beta)\\
            V sin(\alpha)
        \end{bmatrix} - \prescript{B}{}{\textbf{R}_E} * \begin{bmatrix}
            V^w cos(\sigma^w)\\
            V^w sin(\sigma^w)\\
            0
        \end{bmatrix}         
        \\
    \end{aligned}
\end{equation}
\noindent where $\prescript{B}{}{\textbf{R}_E}$ is the Earth to Body rotation matrix \cite{stevens_aircraft_2015}.
In steady planar wind, the aircraft ground velocity vector is:

\begin{equation}
    \begin{aligned}
        \prescript{E}{}{\Vec{V}^g} & = \prescript{E}{}{\Vec{V}^a} + \prescript{E}{}{\Vec{V}^w}\\
        \implies \prescript{E}{}{\Vec{V}^g} &= \prescript{E}{}{\textbf{R}_B} * \prescript{B}{}{\Vec{V}^a} + \prescript{E}{}{\Vec{V}^w}  \\
        \implies \begin{bmatrix}
            V_x\\ 
            V_y \\
            V_z
        \end{bmatrix} & = \prescript{E}{}{\textbf{R}_B} * \begin{bmatrix}
            u \\
            v \\ 
            w 
        \end{bmatrix} + \begin{bmatrix}
            V^w cos(\sigma^w)\\
            V^w sin(\sigma^w)\\
            0
        \end{bmatrix}   
    \end{aligned}
    \label{eq: Ground velocity vector from airspeed and wind speed}
\end{equation}

\noindent where $\prescript{E}{}{\textbf{R}_B} = \prescript{B}{}{\textbf{R}_E}^{-1}=\prescript{B}{}{\textbf{R}_E}^{T}$ is the Body to Earth rotation matrix. For constant altitude flight ($\gamma=0$), ground velocity is given by:

\begin{equation}
    \begin{aligned}
        V_x = V^g cos(\chi)&= V^a cos(\sigma) + V^w cos(\sigma^w); \\ 
        V_y = V^g sin(\chi)&= V^a sin(\sigma) + V^w sin(\sigma^w)
    \end{aligned}
    \label{eq: relationship between airspeed, heading, ground speed and wind speed.}
\end{equation}

In steady wind, conventional aircraft maintain a desired course angle ($\chi$) by "crabbing" into the wind at angle $\sigma_{crab}$. The crab angle can be calculated from airspeed $V^a$, heading $\sigma$, course angle $\chi$, wind speed $V^w$ and wind heading $\sigma_w$ per \cite{pilot_handbook}. However, this method makes a restrictive assumption that the airspeed $V^a$ is always greater than wind speed $V^w$, which does not allow for $\pm 180^\circ$ crab angles. For eVTOL aircraft, hovering in the presence of a steady tailwind requires a $180^\circ$ crab angle to maintain zero sideslip. 
Eq. (\ref{eq: relationship between airspeed, heading, ground speed and wind speed.}) can be generalized. If $V^g, V^w, \chi$ and $\sigma^w$ are known, heading ($\sigma$) and airspeed ($V^a$) are defined by:

\begin{equation}
    \begin{aligned}
        tan(\sigma) &= \frac{V_y - V^w sin(\sigma^w)}{ V_x - V^w cos(\sigma^w))} &&
        \implies \sigma &= atan2 \left( V_y - V^w sin(\sigma^w), V_x - V^w cos(\sigma^w)   \right)\\
    \end{aligned}
    \label{eq: heading calculation from ground and wind velocity under steady wind}
\end{equation}
\begin{equation}
    \begin{aligned}
        V^a &= \sqrt{(V_x - V^w cos(\sigma^w))^2 + (V_y - V^w sin(\sigma^w))^2}
    \end{aligned}
    \label{eq: Airspeed magnitude from ground velocity and wind velocity in steady wind}
\end{equation}
\noindent where $V_x$ and $V_y$ are given by $V^g cos(\chi)$ and $V^g sin(\chi)$ respectively.  Note that $atan2$ is four-quadrant arc tangent.

The next section explores traversal in zero wind conditions ($V^w = 0$). Because $\Vec{V}^a = \Vec{V}^g$ with no wind, we generalize velocity magnitude as "$v$" for simplicity and assume $\chi = \sigma$. 
Section \ref{Section: Optimal traversal in steady wind} then describes how steady wind impacts traversal between hover waypoints.

\section{Optimal Traversal between Hover Waypoint Pairs with No Wind}
\label{Section: Optimal Traversal in No wind}

This section provides a theoretical basis for defining an energy optimal multi-mode flight profile for Lift+Cruise eVTOL aircraft between two hover waypoints at the same altitude in zero wind. 
Hovering at each waypoint allows inter-waypoint flight segments to be optimized independently.
Lift+Cruise aircraft operate in three flight modes: vertical flight for take-off, hover and landing; forward flight that maximizes energy efficiency; and transition between vertical and forward flight modes. 
Flight mode nomenclature is adopted from the QuadPlane \cite{QP_conference_paper}, such that vertical flight mode is referred to as "$Quad$" (subscript "$Q$"), forward / cruise flight mode as "$Plane$" (subscript "$P$"), and transition mode as "$Hybrid$" (subscript "$H$") because it requires a combination of vertical and forward thrust elements.

Below, we first define a general traversal between two hover waypoints, then we discuss energy consumption over the traversal. Next, we present axioms describing energy consumption as a function of airspeed across all three flight modes for Lift+Cruise eVTOL aircraft. Finally, we present a lemma and proof for deriving energy optimal cruise speed over a pair of hover waypoints subject to acceleration limits.

\subsection{General Traversal between Hover Waypoint Pairs with No Wind}

\begin{figure}[ht!]
    \centering
    \includegraphics[width=0.6\linewidth]{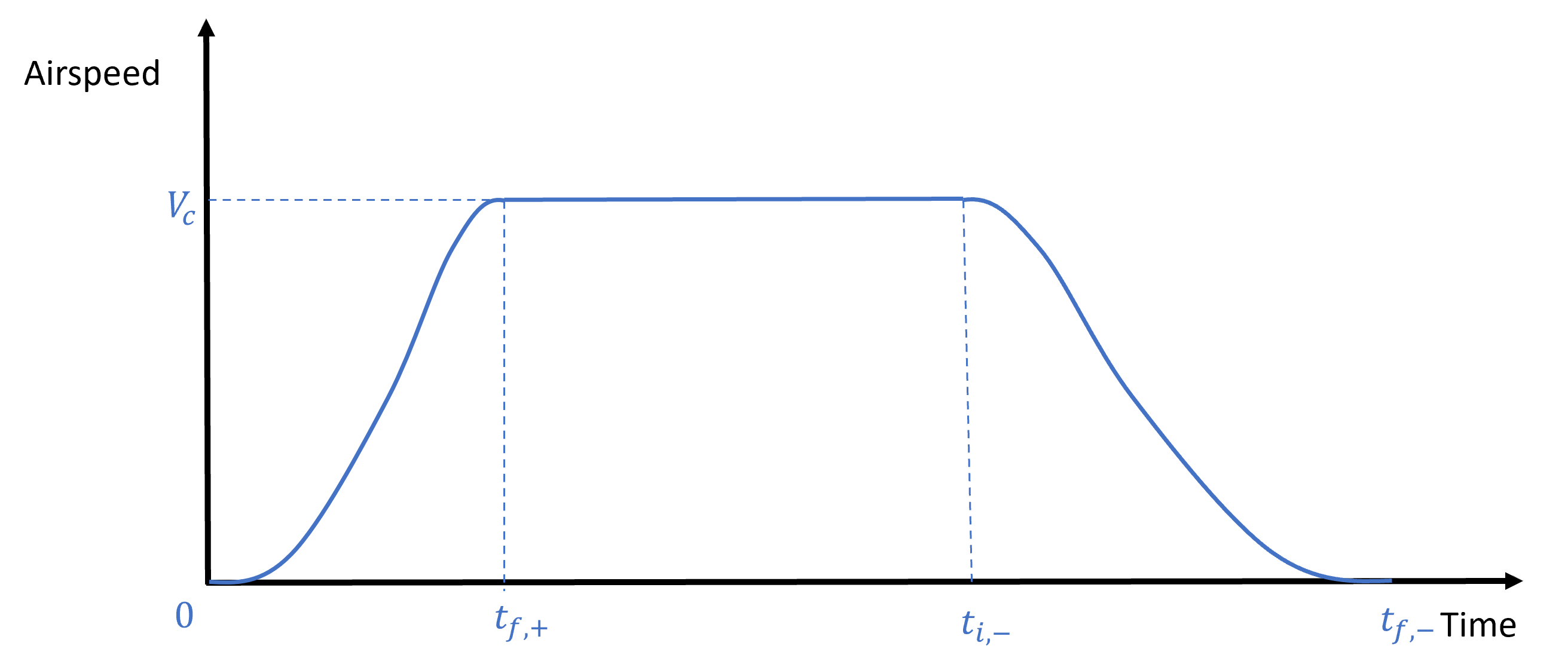}
    \caption{General traversal trajectory $\mathbb{T}_{i,i+1}$ over a segment with hover waypoints.}
    \label{fig: standard stop-n-go trajectory segment}
\end{figure}

Suppose an eVTOL aircraft follows a straight and level trajectory from hover waypoint $\Vec{w}_i = \{x_i, y_i, z_i\}$ to hover waypoint $\Vec{w}_{i+1} = \{x_{i+1}, y_{i+1}, z_{i+1}\}$. With no ambient wind, the hovering aircraft can adjust its course angle at $\Vec{w}_i$ to point directly towards $\Vec{w}_{i+1}$. 
Total path length is given by $l_{i,i+1} = ||\Vec{w}_{i+1} - \Vec{w}_{i}||_{_2}$.
Suppose further that to assure a smooth reference path this trajectory consists of an initial cubic spline accelerated segment from hover at $\Vec{w}_{i}$ to speed $V_c$, an (optional) constant speed segment at $V_c$, and a final cubic spline deceleration segment to hover at $\Vec{w}_{i+1}$. This trajectory is represented in Fig. \ref{fig: standard stop-n-go trajectory segment} and given by:
\begin{equation}
    \begin{aligned}
        \mathbb{T}_{i,i+1} =& \{\mathcal{T}_+, V_c, \mathcal{T}_-\}_{i,i+1}; \\ 
        \mathcal{T}_k = \{V_c, a_{max,k}&, a_{lim,k},V_{QH}, V_{HP}\}
    \end{aligned}
    \label{eq: transition data structure.}
\end{equation}
\noindent where each $\mathcal{T}_k$ defines a cubic spline airspeed profile with subscript $k$ indicating acceleration "$+$" or deceleration "$-$"; $V_c$ indicates the cruise airspeed; $a_{max,k}$ indicates the maximum acceleration or deceleration for the cubic spline; $|a_{lim,k}| \geq |a_{max,k}|$ indicates the aircraft's performance limits for acceleration or deceleration; $V_{QH}$ defines the switching airspeed between $Quad$ and $Hybrid$ mode; and $V_{HP}$ defines the switching airspeed between $Hybrid$ and $Plane$ mode. $V_c$ and $a_{max,k}$ are user defined or optimized by the flight planner; the other variables describe aircraft performance.
Optimal airspeed $V_{HP}$ would typically be close to stall speed since $Plane$ mode is much more efficient that $Hybrid$ mode.
The optimal $Quad$ to $Hybrid$ transition airspeed $V_{QH}$ must ensure stable transition as well as minimizing energy. Flight mode ($FM$) is selected for a given airspeed ($v$) as follows: 
 
\begin{equation}
    FM = \begin{cases}
        "Quad", & \forall 0 \leq v < V_{QH}\\
        "Hybrid", & \forall V_{QH} \leq v < V_{HP}\\
        "Plane", & \forall V_{HP} \leq v \leq V_{lim}
    \end{cases}
    \label{eq: flight mode}
\end{equation}

\begin{figure}[ht!]
\centering
    \begin{tabular}{cc}    
        \includegraphics[width=0.45\linewidth]{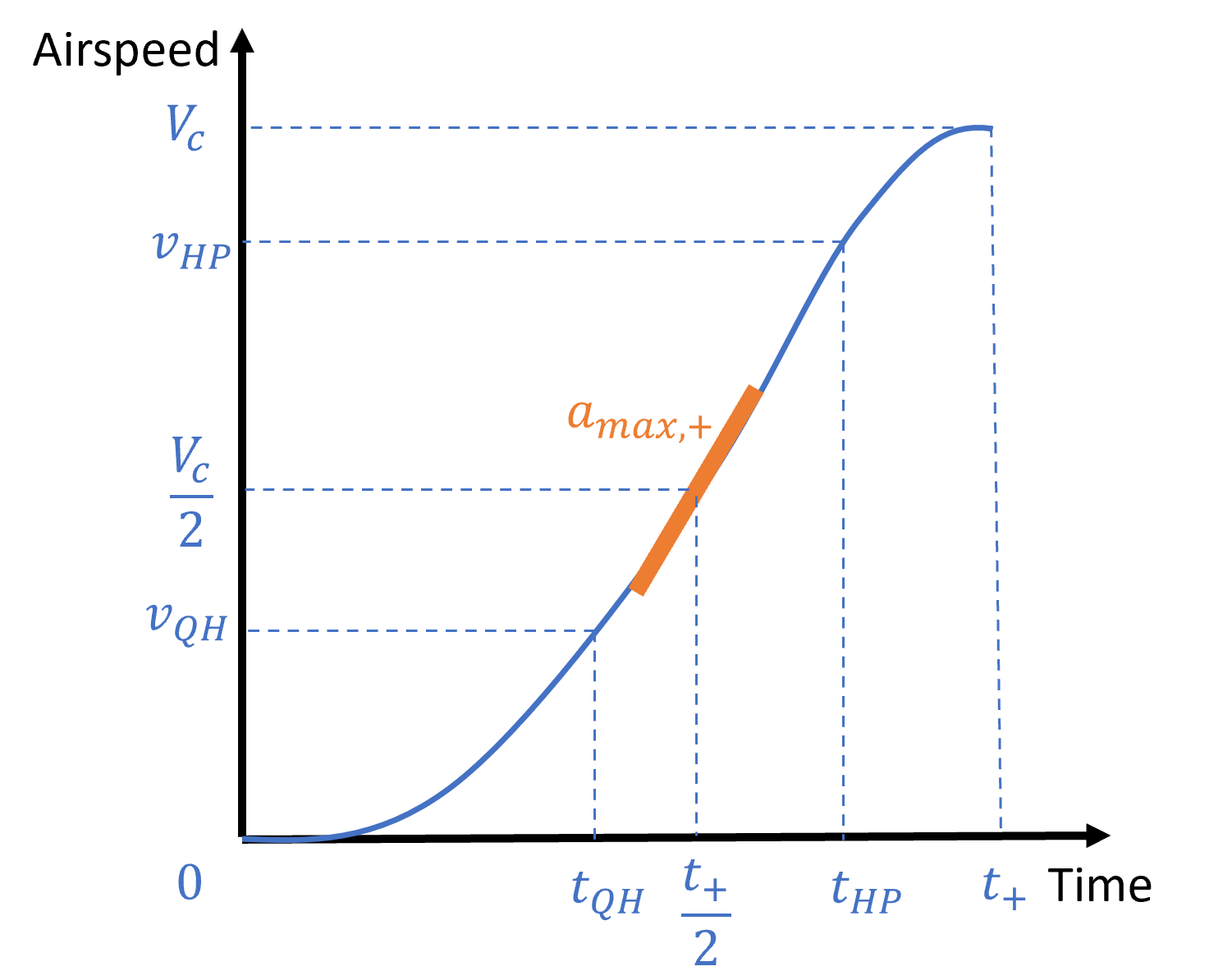} & \includegraphics[width=0.45\linewidth]{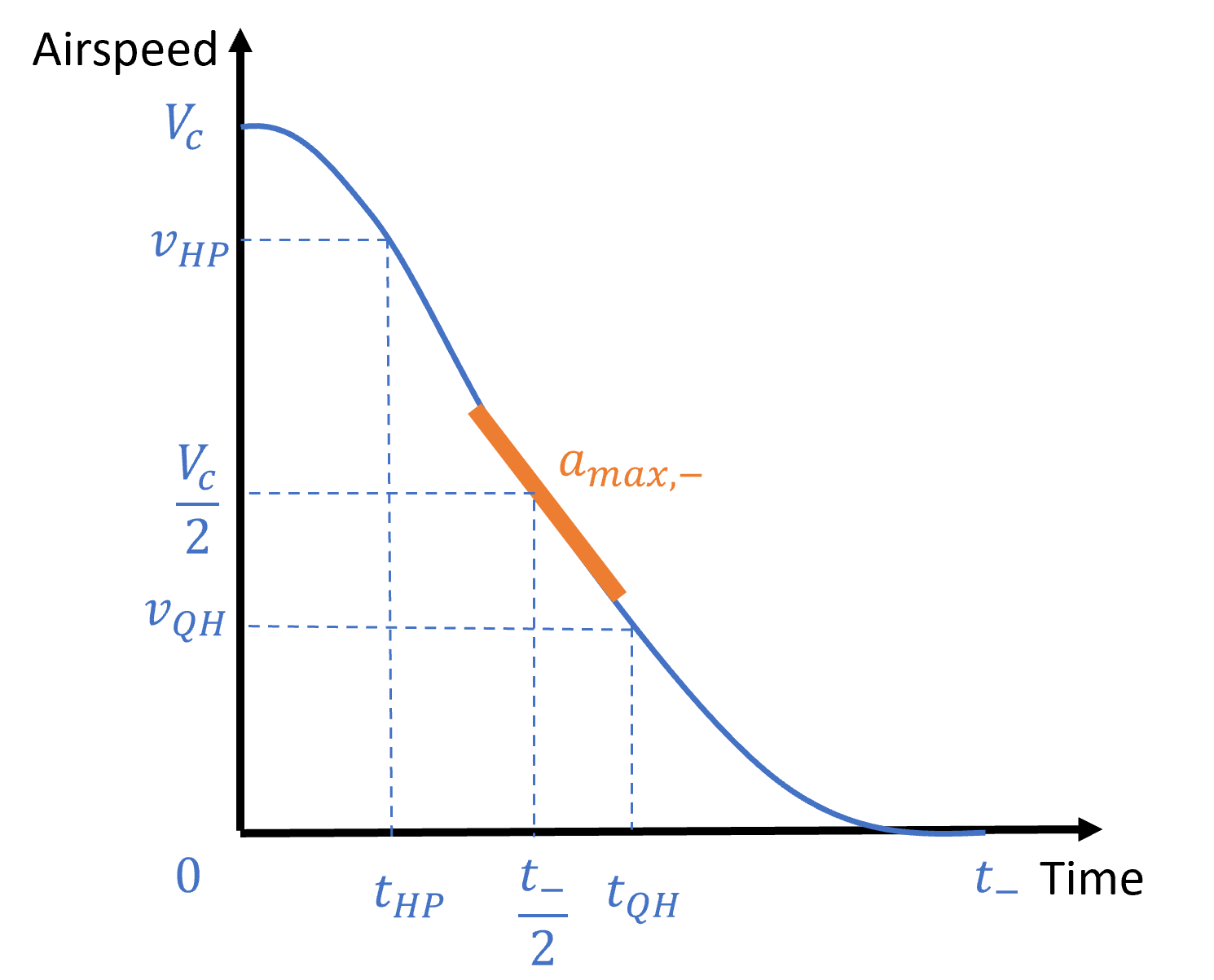}
    \end{tabular}
    \caption{Cubic-spline profile for an airspeed maneuver between hover and $V_c$ defined by $\mathcal{T}_+$ (left) and $\mathcal{T}_-$ (right).}
    \label{fig: Cubic spline}
\end{figure}

Example cubic splines are shown in Fig. \ref{fig: Cubic spline} and follow respective velocity and acceleration profiles given by:
\begin{equation}
\begin{aligned}
    v_k(t) = v_0 + v_1 t + v_2 t^2 + v_3 t^3; &&
    a_k(t) = \frac{d(v_k(t))}{dt}  =& v_1 + 2 v_2 t + 3 v_3 t^2\\
\end{aligned}
\label{eq: cubic spline velocity}
\end{equation}
\noindent subject to the following constraints: 
\begin{equation}
\begin{aligned}
    v_+(t = 0) = 0; v_+(t = t_+) = V_c; &
    a_+(t = 0) = a_+(t = t_+) = 0; &
    a_+(t = \frac{t_+}{2}) = a_{max,+}; & 
    0 < a_{max,+} \leq a_{lim,+}\\
    v_-(t = 0) = V_c; v_+(t = t_-) = 0; &
    a_-(t = 0) = a_-(t = t_-) = 0; &
    a_-(t = \frac{t_-}{2}) = a_{max,-}; &
    a_{lim,-} \leq a_{max,-} < 0
\end{aligned}
\label{eq: constraints on cubic spline velocity}
\end{equation}
\noindent where $t_k$ is the time duration for an accelerated segment given by:

\begin{equation}
\begin{aligned}
    t_+  = (t_{f,+}-0) = \frac{3V_c}{2 a_{max,+}}; && t_- = (t_{f,-} - t_{i,-}) = -\frac{3V_c}{2 a_{max,-}}
\end{aligned}
\label{eq: time during accelreationa nd deceleration}
\end{equation}

\noindent Time at the end of each flight mode ($t_{QH}, t_{HP}$) can be computed using  transition velocities ($v_{QH}, v_{HP}$) in the airspeed cubic spline. Desired acceleration at these times ($a_{QH}, a_{HP}$) can then be computed using acceleration in Eq. (\ref{eq: cubic spline velocity}) for $t = t_{QH}$ and $t = t_{HP}$, respectively. 
Distance $l_k > 0$ traveled over an accelerated segment in zero ambient wind can be calculated by integrating $v_k(t)$ from Eq. (\ref{eq: cubic spline velocity}) over time $t_k$ subject to Eq. (\ref{eq: constraints on cubic spline velocity}) constraints:
\begin{equation}
\begin{aligned}
    l_+ = \frac{3 V_c^2}{4a_{max,+}}; && l_- = -\frac{3 V_c^2}{4a_{max,-}}
\end{aligned}\label{eq: distance traveled under acceleration}
\end{equation}

\noindent The distance traveled over a cruise segment $l_c \geq 0$ for $\mathbb{T}_{i,i+1}$ can then be computed:
\begin{equation}
\begin{aligned}
    l_c = l_{i,i+1} - l_+ - l_- && \implies l_c = l_{i,i+1} - \frac{3 V_c^2}{4}(\frac{1}{a_{max,+}} - \frac{1}{a_{max,-}})
\end{aligned}
    \label{eq: length for a standard stop-n-go segment}
\end{equation} 

For traversal $\mathbb{T}_{i,i+1}$, as cruise speed $V_c$ increases, cruise segment length ($l_c$) decreases. Maximum achievable airspeed ($V_{max}$) is defined when $l_c = 0$.
The minimum segment length required to reach airspeed $V$ is then given by $l_{min}^V = l_+ + l_-$. 
$V_{max}$ is computed for a segment $l_{i,i+1}$ with given $a_{max,+}$ and $a_{max,-}$ as follows:
\begin{equation}
    \begin{aligned}
        l_{i,i+1} = l_+ + l_- &= \frac{3 V_{max}^2}{4} (\frac{1}{a_{max,+}} - \frac{1}{a_{max,-}})\\
        \implies  V_{max} &= \sqrt{\frac{4 l_{i,i+1}}{3} (\frac{1}{a_{max,+}} - \frac{1}{a_{max,-}})^{-1}}
    \end{aligned}
    \label{eq: maximum possible velocity over a segment of length l.}
\end{equation}

\subsection{Energy Cost for General Traversal between Hover Waypoints with No Wind}
Total energy ($E_\mathbb{T}$) for a given traversal $\mathbb{T}_{i,i+1}$ can be computed as the sum of energy consumption over the acceleration, cruise and deceleration phases of flight:
\begin{equation}
    \begin{aligned}
        E_\mathbb{T} = E(\mathcal{T}_+) &+ E(V_c) + E(\mathcal{T}_-);\\
        E(V_c) &= \mathcal{F}(V_c) *l_c 
    \end{aligned}
    \label{eq: Total energy for traversal.}
\end{equation}
\noindent where $\mathcal{F}(V_c) = \frac{E(V_c)}{l_c} = \frac{P(V_c)}{V_c}$ is the energy consumed per distance traveled while flying at constant airspeed $V_c$, and $P(V_c)$ is the power consumed to maintain airspeed $V_c$ in no wind.

The energy consumed during acceleration $E(\mathcal{T}_k)$ is a function of $V_c$ and $a_{max,k}$ as well as fixed parameters $a_{lim,k},V_{QH}$ and $ V_{HP}$. 
Let $V_c^*$ be the most energy efficient cruise airspeed, i.e. $\mathcal{F}(V_c^*) \leq \mathcal{F}(V_c), \forall V_c : (0, V_{lim}]$, where $V_{lim}$ is the vehicle's maximum sustainable airspeed. For eVTOL aircraft with a "Lift+Cruise" configuration \cite{uam_vtol_types}, $V_c^*\geq V_{stall}$ in $Plane$ mode where $V_{stall}$ is aerodynamic stall speed. 

The following axioms describe the characteristics of $E(\mathcal{T}_k)$ and $\mathcal{F}(V_c)$. 
These axioms are confirmed for the QuadPlane in the results section and will generally apply to a "Lift+Cruise" eVTOL design.

\begin{itemize}
    \item \textbf{Axiom 1:} $\mathcal{F}(V_c)$ is a positive definite function that monotonically decreases as $V_c$ increases up to $V_c^*$, i.e. 
    $ \mathcal{F}(V_{c_1})>\mathcal{F}(V_{c_2}),\forall V_{c_1}< V_{c_2} \leq V_c^*$. Also, $\lim_{V_c \to 0} \mathcal{F}(V_c) \to \infty$. 

    \item \textbf{Axiom 2:}
    $\mathcal{F}(V_c)$ monotonically increases as $V_c >  V_c^*$ increases, i.e. $ \mathcal{F}(V_{c_2})>\mathcal{F}(V_{c_1})\forall V_{c}^*<V_{c_1}<V_{c_2}<V_{lim}$. 

    \item \textbf{Axiom 3:} Energy consumed over accelerated flight $E(\mathcal{T}_k)$ is positive definite. $E(\mathcal{T}_k)$ monotonically increases as $V_c$ increases and monotonically decreases as $|a_{max,k}|$ increases due to the corresponding decrease in $t_k$.
    
\end{itemize}

\subsection{Energy Optimal Traversal in No Wind}
Energy consumption for traversal between hover waypoints ($E_{\mathbb{T}}$) given by Eq. (\ref{eq: Total energy for traversal.}) can be optimized over parameters $a_{max,+}$, $V_c$, and $a_{max,-}$.
Define the following constrained multi-variable optimization problem for $E_{\mathbb{T}}$: 
\begin{mini}
    {V_c,a_{max,+},a_{max,-}}{E_{\mathbb{T}} =  E(\mathcal{T}_+)(V_c,a_{max,+}) + \mathcal{F}(V_c)*l_c + E(\mathcal{T}_-)(V_c,a_{max,-})}
    {}{}
    \addConstraint{l_c \geq 0}
    \addConstraint{0 < V_c \leq V_{max}\leq V_{lim}}
    \addConstraint{0 < a_{max,+} \leq a_{lim,+}}
    \addConstraint{a_{lim,-} \leq a_{max,-} < 0}
    \label{eq: minimization problem with constraints}
\end{mini}

Critical points for constrained optimization are found at the corners of the parameter constraint boundary hypercube in addition to the solution(s) of the partial derivatives with respect to each variable \cite{blank2006calculus}. \textit{Solution critical points} could be local minima, local maxima or saddle points. The set of eight \textit{boundary critical points} $B^{cr}$ for the optimization problem in Eq. (\ref{eq: minimization problem with constraints}) are defined by constraints on $V_c$, $a_{max_+}$ and $a_{max_-}$. 
The closed equality constraints directly define boundary critical values. 
For the zero open inequality constraints, define boundary critical values ${V}_{c} \xrightarrow{} 0^+$, $a_{max,+} \xrightarrow{} 0^+$ and $a_{max,-} \xrightarrow{} 0^-$, respectively.
The elements in $B^{cr}$ can be visualized as the corners points for a cuboid with ${V_c}$, $a_{max,+}$ and $a_{max,-}$ as the three axes.
The set of \textit{solution critical points} $S^{cr}$ satisfies $\frac{\partial E_{\mathbb{T}}}{\partial {V_c}}=0$, $\frac{\partial E_{\mathbb{T}}}{\partial a_{max,+}}=0$ and $\frac{\partial E_{\mathbb{T}}}{\partial a_{max,-}}=0$ for a given value set (${V_c}$, $a_{max,+}$, $a_{max,-}$), within the 3D space defined by $B^{cr}$. 
The global minimum is found at one of the \textit{boundary} or \textit{solution critical points} per \cite{thomas_calculus}.

For a given ${V_c}$, $E(\mathcal{T}_k)$ decreases as $|a_{max,k}|$ increases per Axiom 3. However, the corresponding distance traveled $l_k$ under acceleration $a_{max,k}$ decreases per Eq. (\ref{eq: distance traveled under acceleration}), so $l_c$ and $E({V_c}) = \mathcal{F}({V_c})*l_c$ increase. 
Similarly, for a given $a_{max,k}$, $E(\mathcal{T}_k)$ increases as ${V_c}$ increases per Axiom 3. However, $\mathcal{F}({V_c})$ decreases as ${V_c}$ increases for ${V_c}\leq V_c^*$ per Axiom 1, and $l_c$ decreases as ${V_c}$ increases per Eq. (\ref{eq: length for a standard stop-n-go segment}). These trade-offs require analytical solution of  Eq. (\ref{eq: minimization problem with constraints}). 
        
The following Lemma defines energy optimal traversal over a straight and level segment between two hover waypoints subject to aircraft acceleration constraints. 

\begin{lemma}
\label{Lemma: 1}
    Let $\Vec{w}_i$ and $\Vec{w}_{i+1}$ be sequential hover waypoints at the same altitude traversed by trajectory $\mathbb{T}_{i,i+1}$. 
    If $S^{cr}=\emptyset$ and $l_{i,i+1} \leq l_{min}^{V_c^*}$, energy optimal traversal is given by $\mathbb{T}_{i,i+1}^*=\{\mathcal{T}_+^*$, $\mathcal{T}_-^*\}_{i,i+1}$ at boundary critical point $b^{cr} = (V_{max}, a_{lim,+}, a_{lim,-}) \in B^{cr}$.
    If $S^{cr}=\emptyset$ and $l_{i,i+1} > l_{min}^{V_c^*}$, energy optimal traversal is given by $\mathbb{T}_{i,i+1}^* = \{\mathcal{T}_+^*$, $V_c^*$, $\mathcal{T}_-^*\}_{i,i+1}$ at boundary critical point $b^{cr}$.
    If $S^{cr} \neq \emptyset$, let $s^{cr} = (V_c^{cr}, a_{max,+}^{cr}, a_{max,-}^{cr}) \in S^{cr}$ be the solution critical point with minimum $E_{\mathbb{T}}$.  
    If $E_{\mathbb{T}}(s^{cr}) < E_{\mathbb{T}}$($b^{cr})$, optimal traversal is given by $s_{cr}$; otherwise, optimal traversal is given by $b^{cr}$.
\end{lemma}

\renewcommand\qedsymbol{$\blacksquare$}
\begin{proof}
    Total energy consumed $E_{\mathbb{T}}$ for the traversal $\mathbb{T}_{i,i+1}$ is given by Eq. (\ref{eq: Total energy for traversal.}) and is minimized per the multi-variable optimization problem defined in Eq. (\ref{eq: minimization problem with constraints}). The Lemma defines cases categorized by the existence of solution critical point(s) and by segment length. Each case is considered below. 
    \begin{itemize}
        \item  \underline{$S^{cr} = \emptyset$:}
        
        In this case, no solution critical point exists, so the global minima for $E_{\mathbb{T}}$ must be one of the boundary critical points. Consider two sub-cases based on traversal segment length:
        \begin{itemize}
            \item \underline{Short segment:}  $l_{i,i+1} \leq l_{min}^{V_c^*}$
            
            In this case, the maximum achievable airspeed $V_{max}$ is less than or equal to $V_c^*$ due to the short segment length. From Axiom 1, $\mathcal{F}({V_c})$ decreases as ${V_c}$ increases for all ${V_c} \leq V_c^*$. For a given segment length $l_{i,i+1}$, $l_c$ also decreases as ${V_c}$ increases. From Axiom 3,  energy under acceleration $E(\mathcal{T}_k)$ decreases with an increase in $|a_{max,k}|$. To minimize $E_{\mathbb{T}}$, ${V_c}$ and $|a_{max,k}|$ must be maximized. Thus, minimum energy traversal is achieved at the boundary critical point ($V_{max}, a_{lim,+}, a_{lim,-}$) with $l_c= 0$, and the minimum energy traversal is given by $\mathbb{T}_{i,i+1}^*=\{\mathcal{T}_+^*$, $\mathcal{T}_-^*\}_{i,i+1}$, where $\mathcal{T}_k^*$ has ${V_c} = V_{max}$ and $a_{max,k} = a_{lim,k}$.
                
            \item \underline{Long segment:}  $l_{i,i+1} > l_{min}^{V_c^*}$
            
            Segment length does not constrain  $V_{max}$ relative to $ V_c^*$ in this case. Since any cruise airspeed ${V_c} \leq V_{max}$ is possible, two cases are considered separately:
            \begin{itemize}
                \item $V_{lim} \geq {V_c} \geq V_c^*$
                
                From Axiom 2, energy per distance $\mathcal{F}({V_c})$ monotonically increases as ${V_c} \geq V_c^*$ increases. 
                Also, Axiom 3 states that total energy consumed  $E(\mathcal{T}_k^{{V_c}})$ increases as ${V_c}$ increases. 
                Since all terms in Eq. (\ref{eq: Total energy for traversal.}) increase when ${V_c}$ exceeds $V_c^*$, energy use is minimized when ${V_c}= V_c^*$.  
                Also, for a given cruise speed ${V_c}$, an increase in $|a_{max,k}|$ leads to a decrease in total energy $E_{\mathbb{T}}$. Thus, maximizing $|a_{max,k}|$  minimizes $E_{\mathbb{T}}$.

                \item ${V_c} \leq V_c^*$

                Since no solution critical point exists, this case is analogous to the short segment case albeit with a longer cruise segment. Following the same logic as in the short segment case, to minimize $E_{\mathbb{T}}$, one must maximize ${V_c} \leq V_c^*$ and $|a_{max,k}|$. Thus, to minimize energy use ${V_c} = V_c^*$ and $a_{max,k} = a_{lim,k}$.
            \end{itemize}
            
            Combining these two sub-cases, we conclude that cruise at ${V_c}=V_c^*$ with $|a_{max,k}| = |a_{lim,k}|$ minimizes energy use for long segments. Hence, the minimum energy long segment traversal $\mathbb{T}_{i,i+1}^*$ is given by $\{\mathcal{T}_+^*$, $V_c^*$, $\mathcal{T}_-^*\}_{i,i+1}$ such that $\mathcal{T}_k^*$ has ${V_c} = V_c^*$ and $a_{max,k} = a_{lim,k}$.
                
        \end{itemize}

        \item \underline{$S^{cr} \neq \emptyset$:}

        For this case, one or more solution critical points exist within the domain such that all partial derivatives of $E_{\mathbb{T}}$ are zero. 
        Let $(V_c^{cr}, a_{max,+}^{cr}, a_{max,-}^{cr}) \in S^{cr}$ be the solution critical point with minimum energy consumption $E_{\mathbb{T}}$. One of the solution or boundary critical points must minimize $E_{\mathbb{T}}$. The boundary critical point that minimizes $E_\mathbb{T}$ was found for the case where $S^{cr}$ is empty. If $E_\mathbb{T}$ at $(V_c^{cr}, a_{max,+}^{cr}, a_{max,-}^{cr})$ is less than the value at the boundary critical point as found in the previous case. Then, $(V_c^{cr}, a_{max,+}^{cr}, a_{max,-}^{cr})$ is the global minima and the minimum energy traversal is given by $\mathbb{T}_{i,i+1}^* = \{\mathcal{T}_+^*$, $v^{cr}$, $\mathcal{T}_-^*\}_{i,i+1}$, where $\mathcal{T}_k^*$ has ${V_c} = V_c^{cr}$ and $a_{max,k} = a_{max,k}^{cr}$. Otherwise, minimum energy traversal occurs at the boundary critical point given in the previous case.
    \end{itemize}
\end{proof}

To derive a solution to the multi-variable optimization problem in Eq. (\ref{eq: minimization problem with constraints}), aircraft energy consumption data for accelerated and cruise flight are required. With this data, analytical functions can be derived to describe total traversal energy $E_{\mathbb{T}}$ per Eq. (\ref{eq: Total energy for traversal.}) and in turn to solve for the partial derivatives.
Using Lemma \ref{Lemma: 1}, energy optimal traversal between hover waypoints can be found analytically. 
Section \ref{Section: Energy optimal travesal for the QP} analyzes specific QuadPlane vehicle performance data to define traversal energy functions $E_{\mathbb{T}}$. Results show the QuadPlane follows the three Axioms presented above. Lemma \ref{Lemma: 1} is applied to derive QuadPlane velocity and acceleration for energy optimal traversals for different segment lengths.

\section{Traversal between Hover Waypoints in Steady Wind}
\label{Section: Optimal traversal in steady wind}

This section defines level traversal $\mathbb{T}_{i,i+1}^w$ from hover waypoint $\Vec{w}_i = \{x_i, y_i, z_i\}$ to hover waypoint $\Vec{w}_{i+1} = \{x_{i+1}, y_{i+1}, z_{i}\}$ under steady wind of magnitude $V^w$ and heading $\sigma^w$. 
Assume the aircraft must hover at a heading pointed directly into the wind at the beginning and end of each traversal.
Straight-line traversal is the minimum distance solution, requiring an initial acceleration from hover at $\Vec{w}_{i}$ to ground speed $V_c^g$ along course $\chi$ direct to waypoint $\Vec{w}_{i+1}$, an (optional) constant ground speed segment at $V^g_c$, and a final deceleration to hover at $\Vec{w}_{i+1}$. 
Aircraft heading $\sigma$ must be continuously adjusted over the traversal per Eq. (\ref{eq: heading calculation from ground and wind velocity under steady wind}) to maintain desired ground course $\chi$.

\subsection{Straight-line Traversal in Steady Wind}
General traversal in steady wind is given by $\mathbb{T}_{i,i+1}^w = \{\mathcal{T}_+^w, V^g_c, \mathcal{T}_-^w,$\}, where each $\mathcal{T}_k^w$ defines an acceleration profile in steady wind with subscript $k$ indicating acceleration "$+$" or deceleration "$-$". Over this accelerated profile, the ground speed follows a cubic spline with zero initial and final acceleration. Further, the acceleration profile is given by  $\mathcal{T}_k^w = \{V^g_c, a^g_{max,k}, a^a_{lim,k},V^a_{QH}, V^a_{HP},V^w,\sigma^w,\Dot{\sigma}_{lim}\}$, where $V^g_c$ indicates the cruise ground speed for the traversal; $a^g_{max,k}$ indicates maximum ground acceleration or deceleration; $a^a_{lim,k}$ indicates the aircraft's airspeed performance limits; $V^a_{QH}$ defines the transition airspeed between $Quad$ and $Hybrid$ mode; $V^a_{HP}$ defines the transition airspeed between $Hybrid$ and $Plane$ mode; and $\Dot{\sigma}_{lim}$ is the maximum acceptable heading rate over the accelerated segment, characterized by aircraft performance limits.
Straight-line course angle $\chi$ is computed from the waypoint coordinates, i.e. $\chi = tan^{-1} \left( \frac{y_{i+1} - y_i}{x_{i+1} - x_i}\right)$. At any time instant, airspeed $V^a$ and heading $\sigma$ can be computed from ground speed $V^g$ and $\chi$ using Eq. (\ref{eq: heading calculation from ground and wind velocity under steady wind}). Flight mode ($FM$) is selected for a given airspeed $V^a$ based on the transition velocities $V^a_{QH}$ and $V^a_{HP}$. 

In the presence of a steady wind ($V^w, \sigma^w$), an aircraft under acceleration must also constantly adjust its heading $\sigma$ to follow the desired cubic spline ground speed $V^g$. This can be problematic for traversal in a tail wind. To hover in a steady wind, aircraft heading $\sigma$ must point into the wind $\sigma = \sigma^w+180^\circ$ and airspeed must match wind speed $V^a = V^w$. 
Consider traversal in the extreme case of a pure tail wind ($\sigma^w = \chi$). To follow a straight-line path along $\chi$, airspeed magnitude must decrease with an increase in ground speed until $V^g = V^w$. At this point, the aircraft heading $\sigma$ must instantaneously change from pointing into the wind to pointing along the desired course, i.e. $\sigma = \sigma^w+180^\circ \to \sigma^w$. 
Such a step heading change is impossible. As relative wind direction ($\Delta \sigma = \chi - \sigma^w$) increases from zero ($\Delta \sigma =0$ in pure tail wind), the required heading rate of change is not unbounded but still can exceed aircraft performance limit $\Dot{\sigma}_{lim}$. 

The following Lemma specifies when straight-line traversal is possible in steady wind.  In addition to notation already defined, let $a^a_k$ define aircraft acceleration with respect to the relative wind where $k$ denotes acceleration (+) or deceleration (-).

\begin{lemma}
\label{Lemma: Straight Line Traversal in Steady Wind.}
    Suppose an eVTOL aircraft begins and ends a constant altitude waypoint traversal in a hover condition with heading directly into the wind.  
    General traversal $\mathbb{T}_{i,i+1}^w$ with zero sideslip requires the following constraint:
    \begin{equation}
        V^a(t) > V^w sin(|\chi (t) -\sigma^w|)
        \label{eq: airspeed constraint in steady wind.}
    \end{equation}
    Maintaining a constant course angle $\chi = tan^{-1} \left(\frac{y_{i+1}-y_i}{x_{i+1}-x_i}\right)$ path given acceleration profile $a^a_k(t)$ requires 
    \begin{equation}
    \begin{aligned}
        \Dot{\sigma}(t) = \frac{-V^w sin(\chi - \sigma^w)}{V^a(t) \sqrt{{V^a(t)}^2 -{V^w}^2 sin^2(\chi-\sigma^w)}} * a^a_k(t)
    \end{aligned}
        \label{eq: heading rate for straight-line traversal.}
    \end{equation}
    $\mathbb{T}_{i,i+1}^w$ therefore can only follow a constant course angle (straight) path when
    \begin{equation}
    \begin{aligned}
        -\Dot{\sigma}_{lim} \leq \Dot{\sigma}(t) \leq \Dot{\sigma}_{lim} 
    \end{aligned}
        \label{eq: heading rate constraint for straight-line traversal.}
    \end{equation}
    Straight line traversal is not possible when 
    $\Dot{\sigma}_{lim}$ with $|a^a_k(t)| \leq |a^a_{lim,k}|$ cannot be satisfied. 
    In this case, maneuver primitives \cite{maneuver_primitive} that satisfy constraints but do not follow a straight path must be executed at the start and end of $\mathbb{T}_{i,i+1}^w$.
\end{lemma}

\renewcommand\qedsymbol{$\blacksquare$}
\begin{proof}
    Desired heading $\sigma$ to maintain a given course $\chi$ can be derived from  Eq. (\ref{eq: heading calculation from ground and wind velocity under steady wind}). 
    Eq. (\ref{eq: relationship between airspeed, heading, ground speed and wind speed.}) can be used to compute $sin (\sigma - \chi) = \frac{V^w}{V^a} sin(\chi - \sigma^w)$. Since $0 \leq sin (|\sigma - \chi|) \leq1$, it can be concluded that $V^a > V^w sin(|\chi -\sigma^w|)$ for traversal.
    \noindent Further, 
    \begin{equation}
        \sigma = \chi + sin^{-1} \left(\frac{V^w sin(\chi-\sigma^w)}{V^a}\right)
        \label{eq: heading for traversal in steady wind in terms of airspeed.}
    \end{equation}

    \noindent Differentiating Eq. (\ref{eq: heading for traversal in steady wind in terms of airspeed.}) and setting $\Dot{\chi} = 0$ yields the heading rate $\Dot{\sigma}(t)$ for straight-line traversal shown in Eq. (\ref{eq: heading rate for straight-line traversal.}).  
    
    At each hover waypoint, $V_a = V_w$. 
    If $\Dot{\sigma}(t)$ fails to meet the $\Dot{\sigma}_{lim}$ constraint in Eq. (\ref{eq: heading rate constraint for straight-line traversal.}), traversal between the two hover waypoints must allow a change in course angle.
    During the smooth course angle transition, the aircraft cannot follow a straight-line path. A maneuver primitive, by definition, specifies a kinematic transformation between two reference states.  Such a transformation can be constructed to smoothly change heading from its hover value to a designated cruise value without violating kinematic constraints, and vice versa during deceleration to hover.
\end{proof}

In a pure tail wind ($\Delta \sigma = 0^\circ$), if $V^g_c < V_w$,  straight-line travel is always mathematically possible. In this scenario straight-line traversal will take place with the aircraft headed into the wind, appearing to fly "backwards" from the ground observer perspective.
In a pure headwind, i.e. $\Delta \sigma = 180^\circ$, $sin (\Delta \sigma) = 0$ so $\Dot{\sigma}(t) = 0$ per Eq. (\ref{eq: heading rate for straight-line traversal.}); straight-line traversal is always possible for any desired cruise ground speed $V^g_c > 0$ as long as the aircraft can actually progress toward the target waypoint, i.e. $V^g_c + V_w \leq V^a_{lim}$. 

\subsection{Energy Cost for Traversal in Steady Wind}
Total energy consumed $E_{\mathbb{T}}^w$ for traversal $\mathbb{T}_{i,i+1}^w$ is given by the sum of energy consumption over the acceleration, cruise and deceleration phases of flight:
\begin{equation}
    \begin{aligned}
        E_\mathbb{T}^w = E(\mathcal{T}_+^w) &+ E(V^g_c) + E(\mathcal{T}_-^w)
    \end{aligned}
    \label{eq: Total energy for traversal in steady wind.}
\end{equation}
Energy cost for accelerated segments $E(\mathcal{T}_k^w)$ depends on airspeed time history $V^a(T_k)$, where $T_k$ is the time interval for accelerated segment $k$.
$V^a(T_k)$ in turn depends on heading rate $\Dot{\sigma}(T_k)$, desired ground speed profile $V^g(T_k)$, wind speed $V^w$ and direction $\sigma^w$.
Over the cruise segment, energy consumed is characterized by  cruise segment length $l_c$ and cruise ground speed $V^g_c$ in addition to $\chi$, $V^w$ and $\sigma^w$. 
$E_{\mathbb{T}}^w$ can thus be optimized over constrained parameters $\Dot{\sigma}$, $V^g$. 
Define the following constrained multi-variable optimization problem for $E_{\mathbb{T}}$ in steady wind: 
\begin{mini}
    {V^g,\Dot{\sigma}}
    {E_{\mathbb{T}^w} = E(\mathcal{T}_+^w) + \mathcal{F}(V^a_c) * l_c^w + E(\mathcal{T}_-^w)}
    {}{}
    \addConstraint{l_c^w \geq 0}
    \addConstraint{0 < V^g \leq V^g_{max}}
    \addConstraint{0 < V^a \leq V^a_{lim}}
    \addConstraint{a^a_{lim,-} \leq a^a \leq a^a_{lim,+}}
    \addConstraint{-\Dot{\sigma}_{lim} \leq \Dot{\sigma} \leq \Dot{\sigma}_{lim} }
    \addConstraint{0 \leq V^w \leq V^a_{QH}}
    \label{eq: minimization problem with constraints in steady wind}
\end{mini}

The first constraint ensures cruise segment length is non-negative, i.e. the desired cruise ground/airspeed is achievable. $V^g_{max}$ is the maximum ground speed achievable over the segment length. $V^a_{lim}$ is maximum aircraft airspeed. Limits on ground speed are imposed indirectly via airspeed and heading rate constraints. $a^a_{lim,k}$ is the performance limit on airspeed acceleration, and $\Dot{\sigma}_{lim}$ is maximum sustainable heading rate.
The last constraint for the optimization problem restricts wind speed such that the aircraft can operate in Vertical (Quad) mode to hover at each waypoint. The energy optimal traversal solution varies as a function of course angle $\chi$, wind magnitude $V^w$ and wind heading $\sigma^w$ in addition to the variables considered in Lemma \ref{Lemma: 1}.

\subsection{Energy Optimal Straight-line Traversal in Steady Wind}

Let the most efficient cruise airspeed in steady wind be ${V^a_c}^*$.
For straight-line traversal, course angle is computed from the hover waypoint coordinates, i.e. $\chi = tan_{-1} \left( \frac{y_{i+1} - y_i}{x_{i+1}-x_i}\right)$. Aircraft heading $\sigma$ during cruise in steady wind can then be determined using Eq. (\ref{eq: heading for traversal in steady wind in terms of airspeed.}). 
Eq. (\ref{eq: relationship between airspeed, heading, ground speed and wind speed.}) prescribes the corresponding cruise ground speed ${V^g_c}^*$ for straight-line traversal in steady wind with magnitude $V^w$ and direction $\sigma^w$. 
Next, consider traversal $\mathbb{T}_{i,i+1}^w = \{\mathcal{T}_+^w, {V^g_c}^*, \mathcal{T}_-^w,$\} in steady wind, composed of a ground acceleration segment $\mathcal{T}_+^w$, followed by an optional cruise segment at the most energy efficient cruise ground speed ${V^g_c}^*$ and finally a ground deceleration segment $\mathcal{T}_-^w$. 
The accelerated segments are characterized by  $\mathcal{T}_k^w = \{{V^g_c}^*, a^g_{max,k}, a^a_{lim,k},V^a_{QH}, V^a_{HP},V^w,\sigma^w,\Dot{\sigma}_{lim}\}$, where $k$ can be $+$ or $-$ to denote acceleration or deceleration, respectively. 

Algorithm \ref{alg: Straight-line traversal feasibility and computation} describes a \texttt{straight\_traversal\_in\_wind} function which computes and assesses the feasibility of straight-line traversal in steady wind given constraints on acceleration and heading rate.
Inputs to the function include the two hover waypoints $w_i$ and $w_{i+1}$, the most energy efficient cruise ground speed ${V^g_c}^*$, accelerated traversal parameters $\mathcal{T}_k^w$, discretization time step $dt$, minimum allowable ground cruise speed $\prescript{}{min}{V^g_c}$, and minimum value for the maximum acceleration over the cubic spline ground speed profile $\prescript{}{min}{a^g_{max,k}}$ set by the user.
The function outputs logical \texttt{True} or \texttt{False} for straight-line traversal feasibility $STF$. 
It also outputs the straight-line traversal ground speed profile $V^g(T)$ and heading profile $\sigma(T)$ over the time vector $T$ from initial to end time with step size $dt$.  
The function also outputs vectors $T^w$ and $L^w$ listing time duration and distance covered respectively over the acceleration, cruise and deceleration phases of straight-line traversal. Note that airspeed time profile $V^a(T)$ can be computed using $V^g(T)$ and $\sigma(T)$ using Eq. (\ref{eq: relationship between airspeed, heading, ground speed and wind speed.}). When $STF = False$, the other returned parameters are ignored.

\begin{algorithm}[!ht]
    \caption{Straight line traversal in steady wind and its feasibility}
    \label{alg: Straight-line traversal feasibility and computation}
    \begin{algorithmic}
        \State \textbf{function} \texttt{straight\_traversal\_in\_wind}

        \State \textbf{Input: } $\Vec{w}_i$,$\Vec{w}_{i+1}$, ${V^g_c}^*$, $\mathcal{T}_k^w$, $dt$, $\prescript{}{min}{V^g_c}$, $\prescript{}{min}{a^g_{max,k}}$
        
        \State \textbf{Output: } $STF, V^g(T), \sigma(T), T^w, L^w$ 

        \State $l_{i,i+1} = || \Vec{w}_{i+1} - \Vec{w}_{i}||_2$
        
        \State $\chi  = atan2 \left({y_{i+1}-y_i},{x_{i+1}-x_i}\right)$

        \State $V^g_c = {V^g_c}^*$

        \While{$V^g_c \geq \prescript{}{min}{V^g_c}$}
        \State \{$V^g(T_+),\sigma(T_+),t^w_+,l^w_+,T_+,V^g_c$\}=\texttt{comp\_acc\_seg}
        ($V^g_c,a^g_{max,+},\chi,V^w,\sigma^w, dt, a^a_{lim,+}, a^a_{lim,-}, \Dot{\sigma}_{max}, \prescript{}{min}{a^g_{max,+}}$)
        \\ \Comment{calculations for acceleration segment}

        \State \{$V^g(T_-),  \sigma(T_-), t^w_-, l^w_-, T_-, V^g_c$\}=\texttt{comp\_acc\_seg}
        ($V^g_c, a^g_{max,-}, \chi, V^w, \sigma^w, dt, a^a_{lim,+}, a^a_{lim,-}, \Dot{\sigma}_{max}, \prescript{}{min}{a^g_{max,-}}$)
        \\ \Comment{calculations for deceleration segment}

        \State $l_c^w = l_{i,i+1} - l_+^w - l_-^w$
        
        \If{$l_c^w < 0$}
        \Comment{we obtain a negative cruise distance if $V^g_c$ is not achievable}
            \State $V^g_c \gets 0.9*V^g_c$
        \Else
            \State \textbf{break};
        \EndIf
        \EndWhile

        \State $t_c^w = l_c^w/V^g_c$
        \Comment{calculations for cruise segment}
        
        \State $\sigma_c = atan2 (V^g_c sin(\chi) - V^w sin(\sigma^w), V^g_c cos(\chi) - V^w cos(\sigma^w))$

        \State $T_c = t_{+}^w : dt : t_{+}^w + t_c^w$
        
        \State  $V^g(T_c) = ones(length(T_c),1)*{V^g_c}$

        \State $\sigma(T_c) = ones((length(T_c),1)* \sigma_c$

        \State $T^w = [t_+^w;t_c^w;t_-^w]$
        \Comment{concatenating time periods for acceleration, cruise and deceleration}
        
        \State $L^w = [l_+^w;l_c^w;l_-^w]$
        \Comment{concatenating distance traveled over acceleration, cruise and deceleration}

        \State $T_- \gets T_- + t_+^w + t_c^w$
        \Comment{shifting deceleration time to end of traversal}
        
        \State $T = [T_+; T_c; T_-]$
        \Comment{complete traversal time vector}

        \State $V^g(T) = [V^g(T_+);V^g(T_c);V^g(T_-)]$
        
        \State $\sigma(T) = [\sigma(T_+); \sigma(T_c); \sigma(T_-)]$
        
        \State Compute $\Dot{\sigma}(T) = \frac{d}{dt} \sigma(T)$ 
        
        \If{$|\Dot{\sigma}(t)| \leq \Dot{\sigma}_{lim}\forall t\in T $}
            \State $STF = True$
        \Else 
            \State $STF = False$        
        \EndIf
    \end{algorithmic}
\end{algorithm}

\begin{algorithm}[!ht]
    \caption{Construction of a straight-line accelerated segment in steady wind, constrained by vehicle limits}
    \label{alg: constrained computations for the accelerated segments}
    \begin{algorithmic}
        \State \textbf{function} \texttt{comp\_acc\_seg}
        \State \textbf{Input:} $V^g_c, a^g_{max,k}, \chi, V^w, \sigma^w, dt, a^a_{lim,+}, a^a_{lim,+}, \Dot{\sigma}_{max}, \prescript{}{min}{a^g_{max,k}}$ 
        \State \textbf{Output} $V^g(T_k),  \sigma(T_k), t^w_k, l^w_k, T_k$
        \While{$|a^g_{max,k}| \geq \prescript{}{min}{a^g_{max,k}}$}
            \State $t_k^w = \begin{cases}
                {3{V^g_c}}/{(2 a^g_{max,k})}, & \text{if $k$ is "+"}\\
                -{3{V^g_c}}/{(2 a^g_{max,k})}, & \text{if $k$ is "-"}
            \end{cases}$ \Comment{recall that $a^g_{max,-}$ is  negative}
            \State $l_k^w = \begin{cases}
                {3 {{V^g_c}}^2}/{(4a^g_{max,+})}, & \text{if $k$ is "+"}\\
                -{3 {{V^g_c}}^2}/{(4a^g_{max,+})}, & \text{if $k$ is "-"}
            \end{cases}$
            \State $T_k = 0: dt: t_k^w$
            \State  $V^g(T_k) = \begin{cases}
                cubic (0, {V^g_c}, t_k^w, a^g_{max,k}), & \text{if $k$ is "+"}\\
                cubic ({V^g_c}, 0, t_k^w, a^g_{max,k}), & \text{if $k$ is "-"}
            \end{cases}$
            \State $\sigma(T_k) = atan2 (V^g(T_k) sin(\chi) - V^w sin(\sigma^w), V^g(T_k)cos(\chi) - V^w cos(\sigma^w))$
            \Comment{ using Eq. \ref{eq: heading calculation from ground and wind velocity under steady wind}}
            \State $V^a(T_k) = \sqrt{(V^g(T_k) cos(\chi) - V^w cos(\sigma^w))^2 + (V^g(T_k) sin (\chi) - V^w sin(\sigma^w))^2}$        \Comment{using Eq. (\ref{eq: Airspeed magnitude from ground velocity and wind velocity in steady wind})}
            \State $a^a (T_k) = \frac{d}{dt} V^a(T_k)$
            \State $\Dot{\sigma} (T_k) = \frac{d}{dt} \sigma(T_k)$
            \If{$\{a^a_{lim,-} \leq |a^a(t)| \leq a^a_{lim,+}\} \textbf{\&} \{|\Dot{\sigma} (t)| \leq \Dot{\sigma}_{max}\} \forall t \in T_k$} 
                \State \textbf{break};
            \Else
                \State $a^g_{max,k} \gets 0.9*a^g_{max,k}$
            \EndIf
            
        \EndWhile    
    \end{algorithmic}

\end{algorithm}

The algorithm implicitly uses Lemma \ref{Lemma: 1} to define cruise at the most efficient airspeed if possible while maximizing acceleration and deceleration within constraints. Straight-line distance $l_{i,i+1}$ and course angle $\chi$ are computed, and  cruise ground speed ${V^g_c}^*$ is set. 
Next, the function \texttt{comp\_acc\_seg} obtains time duration $t_+^w$, distance traveled $l_+^w$, ground speed time profile $V^g(T_+)$ and heading profile $\sigma (T_+)$ over the acceleration segment as described in Algorithm \ref{alg: constrained computations for the accelerated segments}. 
Computation for $V^g(T_+)$ uses a \texttt{cubic} function, which takes the initial and final ground speeds along with the time duration and maximum ground acceleration as inputs to compute a cubic-spline speed time profile with zero initial and final acceleration. 
The heading profile $\sigma(T_+)$ over the acceleration segment is computed using the ground speed profile along with wind speed and magnitude per Eq. \ref{eq: heading calculation from ground and wind velocity under steady wind}. 
Airspeed profile $V^a(T_+)$ is derived from Eq. \ref{eq: Airspeed magnitude from ground velocity and wind velocity in steady wind}. 
Airspeed and heading profiles are differentiated with respect to time to define $a^a(T_+)$ and heading rate $\Dot{\sigma}(T_+)$ profiles. 
Constraints on acceleration $a^a_{lim,k}$ and heading rate $\Dot{\sigma}_{max}$ are checked for each time instant $t\in T_+$. If a constraint is not met for any time step, the maximum ground acceleration $a^g_{max,+}$ is reduced by $10\%$, a value that can be tuned by the user.
Parameters for the acceleration segment $t_+^w$, $l_+^w$, $V^g(T_+)$ and $\sigma(T_+)$ are computed recursively until the acceleration and heading rate constraints are met or the maximum ground acceleration over the cubic spline is reduced to $\prescript{}{min}{a^g_{max,+}}$.

The function \texttt{comp\_acc\_seg} is used again for the deceleration segment, where the time period $t_-^w$, distance traveled $l_-^w$, ground speed $V^g(T_-)$ and heading $\sigma(T_-)$  profiles are recursively computed with a $10\%$ reduction in $a^g_{max,-}$, until either the acceleration and heading rate constraints are met for all time instances during the deceleration phase or if the ground deceleration reaches its minimum value.
Next, the cruise distance $l_c^w$ is computed using $l_{i,i+1}$, $l_+^w$ and $l_-^w$. A negative computed value of $l_c^w$ indicates that the segment is not sufficiently long to achieve the desired ground cruise speed ${V^g_c}^*$ at the given acceleration and deceleration values. The acceleration and deceleration profile computation loop is executed recursively with the desired cruise ground speed reduced by $10\%$ until a non-negative value for $l_c^w$ is returned or ground cruise speed reaches its minimum value $\prescript{}{min}{V^g_c}$.

Next, cruise duration $t_c$ is computed and desired ground speed and heading values are assigned for cruise segment $V^g(T_c)$ to obtain $\sigma(T_c)$. Time histories are assembled by concatenating acceleration, cruise and deceleration profiles. Heading rate profile $\Dot{\sigma}(T)$ is computed by differentiating $\sigma(T)$ with respect to time. If the heading rate constraint is not met at all time steps $t\in T$, the algorithm sets $STF$ to \texttt{False}. Otherwise, $STF$ is \texttt{True}. $V^g(T)$, heading profile $\sigma(T), T^w$ and $L^w$ are returned. 
Airspeed $V^a(T)$ and heading can be computed from $V^g(T)$ and $\sigma(T)$ per Eq. (\ref{eq: relationship between airspeed, heading, ground speed and wind speed.}) and input as guidance commands to the aircraft controller.

If the algorithm returns $STF=False$, maneuver primitives must be added at the start and end of traversal and connected with a straight-line cruise segment per Lemma \ref{Lemma: Straight Line Traversal in Steady Wind.}. An optimal maneuver primitive pair could be specified with Optimal Control \cite{optimal_control}, or a simple maneuver pair can be prescribed analytically. 
This work defines a simple cubic spline maneuver primitive for each accelerated segment to complete the algorithm for all steady wind headings. 
Section \ref{Section: QP case study} presents case study results that illustrate cubic spline maneuver primitive design when straight-line traversal is not possible for the QuadPlane small UAS \cite{QP_conference_paper} reference platform.

\subsection{Steady Wind Traversal with Cubic Spline Maneuver Primitives}
\label{subsection: CS maneuver primitive}

This sub-section describes a cubic-spline maneuver primitive over course angle and ground speed to smoothly transition from $V^g = 0$ (at the hover waypoint) to cruise speed $V^g_c$ in steady winds subject to acceleration $a^a_{lim,k}$ and heading rate $\Dot{\sigma}_{max}$ constraints. 
First, define wind angle $\delta$ for straight-line traversal as $\delta = \chi_{st} - \sigma^w$ where $\chi_{st} = atan2 \left({y_{i+1}-y_i},{x_{i+1}-x_i}\right)$.
Aircraft course angle while hovering ($\chi_{h}$) is initialized to match heading into the wind: 
\begin{equation}
    \chi_{h} = \sigma_{h} = \begin{cases}
        \sigma^w + \pi, & \text{if $\delta > 0$} \\
        \sigma^w - \pi, & \text{if $\delta \leq 0$}  
    \end{cases}
\end{equation}

At the end of the acceleration cubic spline maneuver primitive, the ground speed and course angle must match their cruise values. Initially, assume the cruise course angle to be the same as the straight-line course angle, i.e. $\chi_c = \chi_{st}$. Next, compute the time periods for cubic spline ground speed acceleration $t^g_+$ and change in course angle $t^\chi_+$ over the acceleration maneuver primitive subject to maximum ground acceleration $a^g_{max,+}$ and course angle rate $\Dot{\chi}_{max}$ constraints. The acceleration maneuver time $t_+$ is set to the longer time period for ground speed and course angle change:  $t_+ = max(t^g_+, t^\chi_+)$. The time interval for the acceleration maneuver is defined by $T_+ = 0:dt:t_+$. 
Ground speed and course profiles over the acceleration segment for all $t\in T_+$ are prescribed by:
\begin{equation}
\begin{aligned}
    V^g(t) = \begin{cases}
        V^g_0 + V^g_1 t + V^g_2 t^2 + V^g_3 t^3, & \text{if $t\leq t^g_+$}\\
        V^g_c, & \text{if $t^g_+ < t \leq t_+$}
    \end{cases}; && 
    a^g(t) = \frac{d(V^g_k(t))}{dt}  =& V^g_1 + 2 V^g_2 t + 3 V^g_3 t^2\\
    \chi(t) = \begin{cases}
        \chi_0 + \chi_1 t + \chi_2 t^2 + \chi_3 t^3, & \text{if $t\leq t^\chi_+$}\\
        \chi_c, & \text{if $t^\chi_+ < t \leq t_+$}
    \end{cases}; && 
    \Dot{\chi} (t) = \frac{d(\chi_k(t))}{dt}  =& \chi_1 + 2 \chi_2 t + 3 \chi_3 t^2
\end{aligned}
\label{eq: cubic spline under acceleration}
\end{equation}
\noindent subject to the following constraints: 
\begin{equation}
\begin{aligned}
    V^g(t = 0) = 0; && V^g(t = t^g_+) = V^g_c; &&
    a^g(t = 0) = a^g(t = t^g_+) = 0; &&
    a^g(t = \frac{t^g_+}{2}) = a^g_{max,+}\\
    \chi(t = 0) = 0; && \chi(t = t^\chi_+) = \chi_c; &&
    \Dot{\chi}(t = 0) = \Dot{\chi}(t = t^\chi_+) = 0; &&
    \Dot{\chi}(t = \frac{t^\chi_+}{2}) = \Dot{\chi}_{max}
\end{aligned}
\label{eq: constraints on cubic spline in accelration}
\end{equation}
The eight cubic spline coefficients for ground speed and course angle are computed from the above constraints.
Using the ground speed $V^g(T_+)$ and course angle $\chi(T_+)$ time profiles, airspeed $V^a(T_+)$ and heading $\sigma(T_+)$ profiles are computed from Eqs. (\ref{eq: heading calculation from ground and wind velocity under steady wind}) and (\ref{eq: Airspeed magnitude from ground velocity and wind velocity in steady wind}).
Displacement over the accelerated maneuver ($\Delta x_+, \Delta y_+$) is computed by integrating Eq. (\ref{eq: relationship between airspeed, heading, ground speed and wind speed.}) per:
\begin{equation}
\begin{aligned}
    \Delta x_+ = \int_0^{t_+} V^a(t)*cos(\sigma(t)) + V^w*cos(\sigma^w)  dt; && \Delta y_+ = \int_0^{t_+} V^a(t)*sin(\sigma(t)) + V^w*sin(\sigma^w)  dt
\end{aligned}
\label{eq: displacement over the accelerated segment in steady wind.}
\end{equation}

\begin{figure}[!ht]
    \centering
    \includegraphics[width=01.0\linewidth]{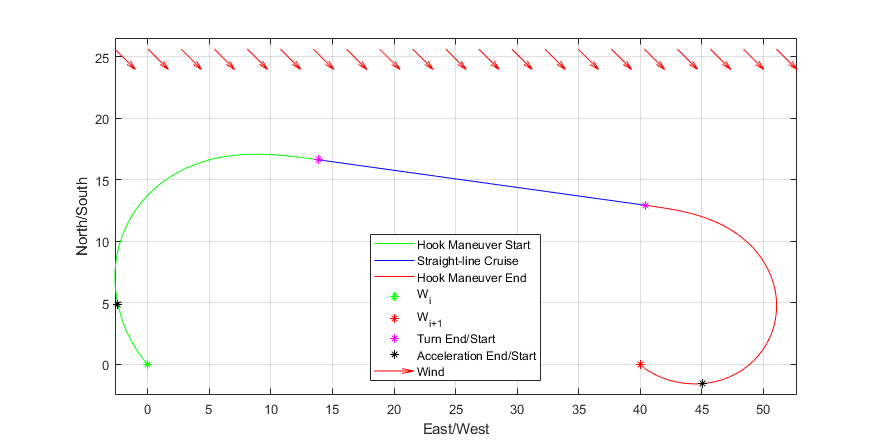}
    \caption {Maneuver primitive traversal in steady wind.}
    \label{fig: Example maneuver primitive}
\end{figure}

Similarly, ground speed $V^g(T_-)$, course angle $\chi(T_-)$, airspeed $V^a(T_-)$ and heading $\sigma(T_-)$ profiles can be computed as well as displacement over the deceleration maneuver $\Delta x_-, \Delta y_-$. 
Next, waypoint coordinates are computed at the end of the acceleration maneuver $\Vec{w}_{m_1} = \Vec{w}_i +  \{\Delta x_+, \Delta y_+, 0\}$ and at the start of the deceleration maneuver $\Vec{w}_{m_2} = \Vec{w}_{i+1} -  \{\Delta x_-, \Delta y_-, 0\}$. The straight-line cruise course angle between the two intermediate maneuver waypoints is computed with $\chi_c^m = atan2 (\Vec{w}_{m_2}(2) - \Vec{w}_{m_1}(2), \Vec{w}_{m_2}(1) - \Vec{w}_{m_1}(1))$. Error in course angle $\Delta \chi = \chi_c^m - \chi_c$ is computed, the cruise course angle $\chi_c$ is set to $\chi_c^m$ and the process iterates until $\Delta \chi$ is less than a small threshold $\epsilon$. 

 heading rate constraint $\Dot{\sigma}(T_k) \leq \Dot{\sigma}_{max}, \forall t\in \{T_+, T_-\}$ is not satisfied,  maximum course angle rate $\Dot{\chi}_{max}$ is reduced by a user-specified percentage until the heading rate constraint is met. Cruise is a straight segment at ground speed $V_c^g$, course angle $\chi_c^m$, airspeed $V^a_c$ and heading $\sigma_c$ which connects the two maneuvers. Fig. \ref{fig: Example maneuver primitive} shows a cubic spline maneuver primitive pair for traversal from hover waypoint $W_i$ (green star) to hover waypoint $W_{i+1}$ (red star) for the QuadPlane sUAS described below with $V_w = 1m/s$, $\sigma^w = 135^\circ$, ${V^a_c} = 2m/s$, $a^g_{max,+} = 1m/s^2$, $a^g_{max,-} = -1m/s^2$ and $\Dot{\sigma}_{max} = 35^\circ$.

\section{QuadPlane Traversal Case Study}
\label{Section: QP case study}

The eVTOL QuadPlane duced in \cite{QP_conference_paper} has four vertical thrust motors, one forward thrust motor, a conventional fixed-wing aircraft design, and standard aircraft control surfaces. It can operate in $Quad$ (vertical), $Plane$ (cruise) and sustained $Hybrid$ (transition) flight modes. QuadPlane performance was characterized in \cite{QP_arXiv_journal_paper} and \cite{QP_JOA_engineering_note}.
This section first uses  published data to model QuadPlane power and energy consumption under steady and accelerated flight.
Next, optimal traversals are presented in no wind conditions, followed by traversals in steady wind. Steady wind constraints for straight-line traversal are characterized.

\subsection{Modeling QuadPlane Power and Energy Consumption}

This section defines the QuadPlane analytical functions required by Lemma \ref{Lemma: 1} to compute total energy for traversal in the presence ($E_\mathbb{T}^w$) and absence ($E_\mathbb{T}$) of steady wind. Data from \cite{QP_JOA_engineering_note} and \cite{QP_arXiv_journal_paper} is utilized to derive curve and surface fits for power and energy consumption over the acceleration, cruise and deceleration traversal phases. 
Power consumption can be computed for any airspeed and acceleration condition. However, energy per unit distance is a function of wind magnitude and direction as well as aircraft ground traversal profile.  

\subsubsection{Power consumption in steady flight}

To calculate total energy consumed over traversal $\mathbb{T}_{i,i+1}^w$ in steady wind, first compute energy consumed over accelerated flight $E(\mathcal{T}_k^w)$ and over the cruise segment $E(V^g_c)$. 
QuadPlane power consumption as a function of airspeed in $Quad$, $Hybrid$ and $Plane$ modes for steady, level flight is derived from \cite{QP_arXiv_journal_paper}.
Power consumption data and corresponding curve fits are shown in Fig. \ref{fig: Power consumption all modes.} with curve fit equations and parameters given in the Appendix. 
Power consumption $P(V^a_c)$ varies as a quartic function of cruise airspeed $V^a_c$  for $Quad$ and $Hybrid$ modes and as a cubic function for $Plane$ mode. 
These curve fits can be used to determine cruise power consumption in any flight mode as a function of cruise airspeed $V^a_c$. Cruise energy is then computed as $E(V^g_c) = P(V^a_c) * t_c^w$ where $t_c^w$ is cruise time for the traversal. Cruise ground speed $V^g_c$ is computed from Eq. (\ref{eq: relationship between airspeed, heading, ground speed and wind speed.}).

\begin{figure}[!ht]
    \centering
    \includegraphics[width=1.0\linewidth]{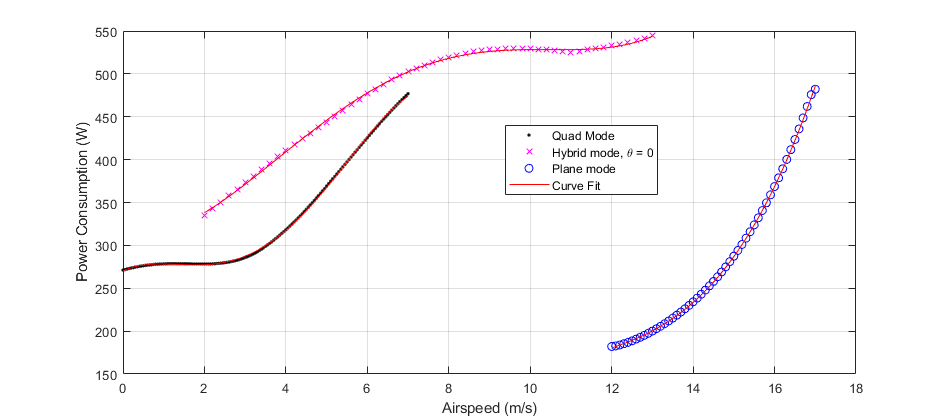}
    \caption{QuadPlane Power consumption vs airspeed for steady level flight in $Quad$, $Hybrid$ and $Plane$ Modes.}
    \label{fig: Power consumption all modes.}
\end{figure}

\subsubsection{Power consumption in accelerated flight}

\begin{figure}[!ht]
    \centering
    \begin{tabular}{cc}
        \includegraphics[width=0.45\linewidth]{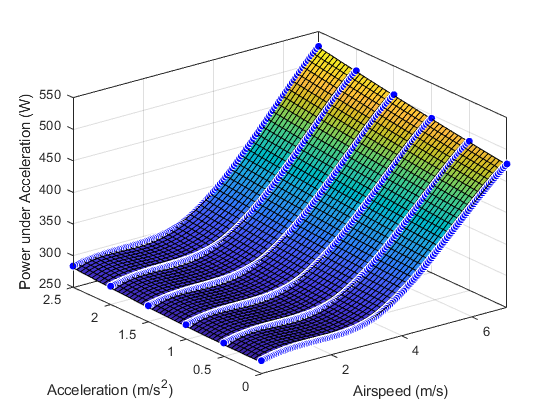} &
        \includegraphics[width=0.45\linewidth]{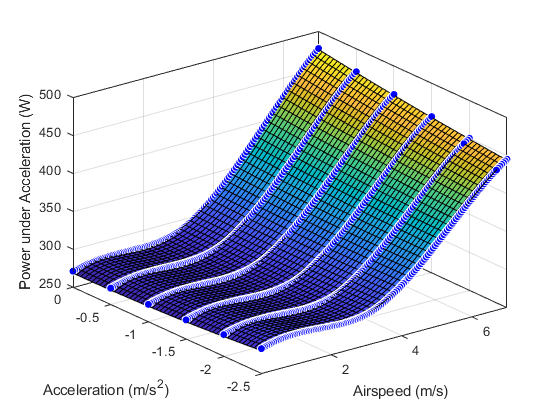}
    \end{tabular}    
    \caption{QuadPlane Power vs Airspeed under Acceleration (left) and Deceleration (right) in $Quad$ Mode. }
    \label{fig: Power under acceleration in $Quad$ mode}
\end{figure}

\begin{figure}[!ht]
    \centering
    \begin{tabular}{cc}
        \includegraphics[width=0.45\linewidth]{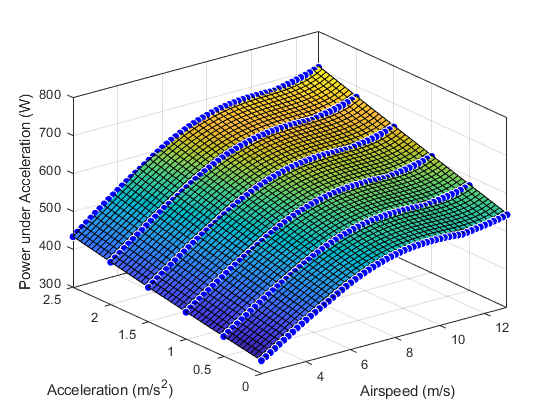} & 
        \includegraphics[width=0.45\linewidth]{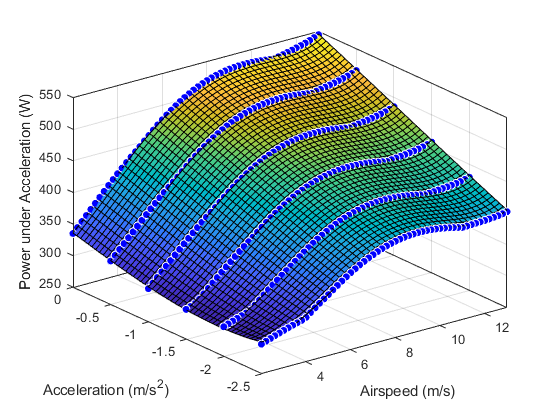}
    \end{tabular}    
    \caption{QuadPlane Power vs Airspeed under Acceleration (left) and Deceleration (right) in $Hybrid$ Mode. }
    \label{fig: Power under acceleration in $Hybrid$ mode}
\end{figure}

Accelerated flight for transitions between hover and cruise requires $Quad$ and $Hybrid$ flight modes. Power consumption in $Quad$ and $Hybrid$ modes over a range of instantaneous acceleration values is computed as a function of airspeed, using force balance and propulsion system models from \cite{QP_arXiv_journal_paper}. Surface fits in Fig. \ref{fig: Power under acceleration in $Quad$ mode} and \ref{fig: Power under acceleration in $Hybrid$ mode} depict a "$poly 53$" surface with specific values shown in the Appendix representing power consumption as a function of airspeed and instantaneous acceleration or deceleration for $Quad$ and $Hybrid$ modes. 
The surface fit for each flight mode considers instantaneous acceleration values from $-2.5m/s^2$ to $2.5m/s^2$ to realize the full range of sUAS QuadPlane performance.

The applicable flight mode for any airspeed $V^a$ can be determined using Eq. \ref{eq: flight mode}, where $V^a_{QH} = 2m/s$ and $V^a_{HP} = 12m/s$ for the QuadPlane. $V^a_{HP}$ is set to be just above QuadPlane stall speed ($11.9 m/s$) above which $Plane$ mode is feasible and more efficient than $Hybrid$ mode. 
The QuadPlane pitches forward as a function of airspeed in $Quad$ mode but maintains zero pitch in $Hybrid$ mode. Thus, a higher $V^a_{QH}$ requires a larger step change in pitch angle. 
$V^a_{QH}$ is set to an airspeed that balances stability and energy consumption considerations. 
rgy consumed over each accelerated segment $\mathcal{T}_k^w$ from Fig. \ref{fig: Cubic spline} can then be computed: $E(\mathcal{T}_k^w) = \int^{t_k^w} P(\mathcal{T}_k^w)dt$. Eq. (\ref{eq: Total energy for traversal in steady wind.}) then defines total energy consumption ($E_\mathbb{T}^w$) over traversal $\mathbb{T}_{i,i+1}$. 
Power consumed under acceleration in $Plane$ mode is neglected because ${V^a_c}^* \simeq V^a_{HP}$ for the QuadPlane.

\subsubsection{Energy per Distance in steady flight with no wind}

\begin{figure}[!ht]
    \centering
    \includegraphics[width=0.9\linewidth]{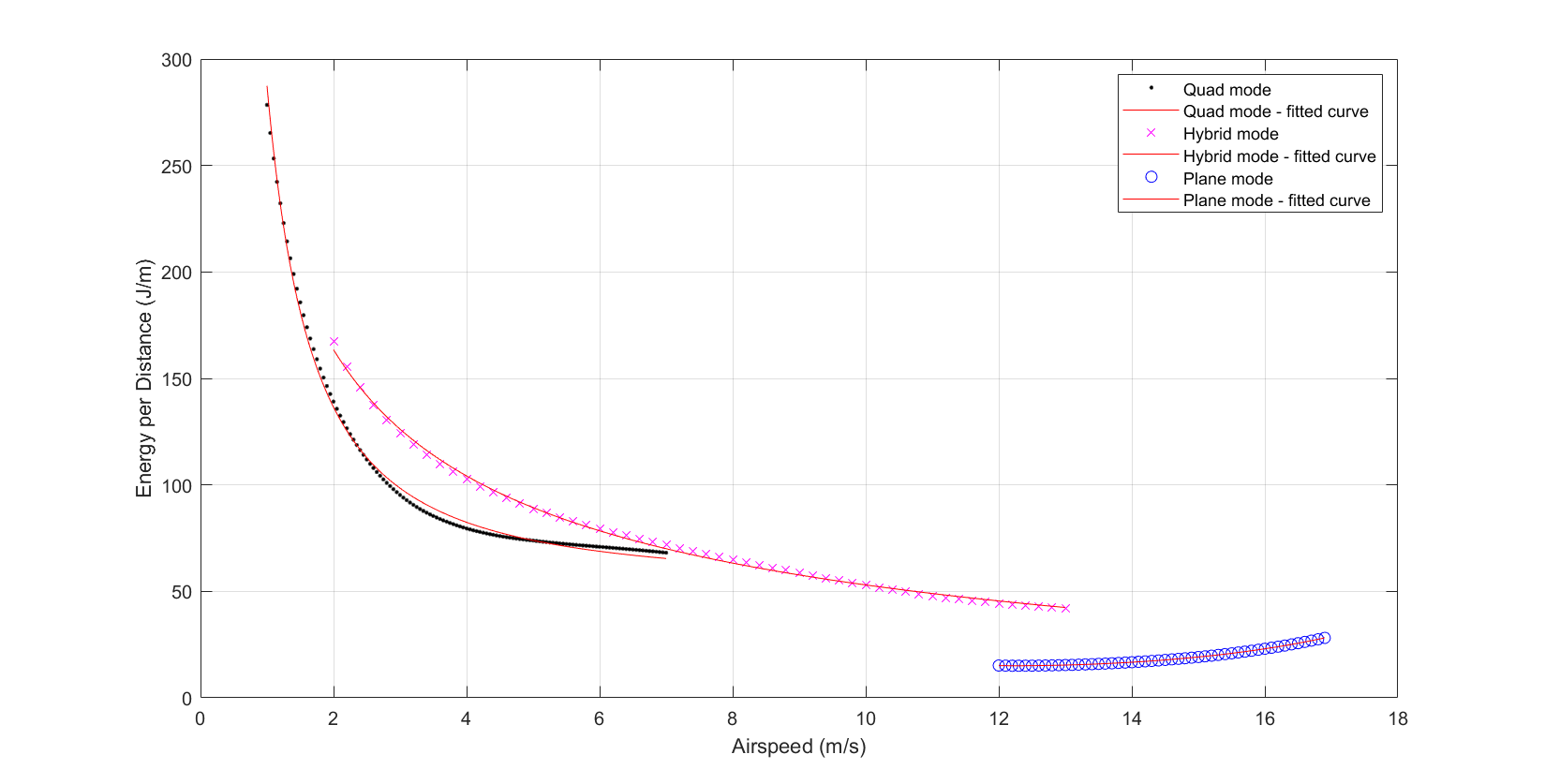}
    \caption{Energy per Distance ($\mathcal{F}$) data and curve fits for the three QuadPlane flight modes.}
    \label{fig: EPD curve fits}
\end{figure}

Energy per Distance ($\mathcal{F} = \frac{P(V_c)}{V_c}$) is computed for speeds above $V_c\geq 1m/s$ in $Quad$ mode since it is impractical from an energy consideration to cruise at a lower speed. $\mathcal{F}$ for $Hybrid$ mode is computed for airspeeds satisfying $V^a_{QH}\leq V_c \leq V^a_{HP}$ since cruise would occur in $Hybrid$ mode only within this airspeed range per Eq. \ref{eq: flight mode}. For $Plane$ mode, $\mathcal{F}$ is computed for $V_{HP}\leq V_c\leq V_{lim}$, where the maximum achievable airspeed for the QuadPlane is $V_{lim} = 16.9m/s$. 
Figure \ref{fig: EPD curve fits} shows QuadPlane energy per distance $\mathcal{F}$ data and curve fits with parameters  listed in the Appendix. 
Fig. \ref{fig: EPD curve fits} confirms that $\mathcal{F}$ for the QuadPlane supports Axioms 1 and 2 for all flight modes.

\subsubsection{Energy consumption over accelerated flight with no wind}

\begin{figure}[ht!]
\centering
    \begin{tabular}{cc}    
        \includegraphics[width=0.45\linewidth]{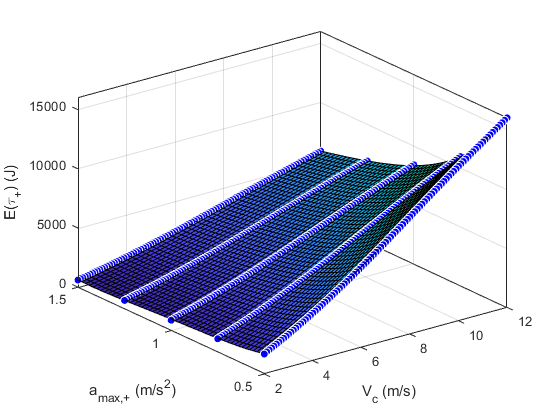} & \includegraphics[width=0.45\linewidth]{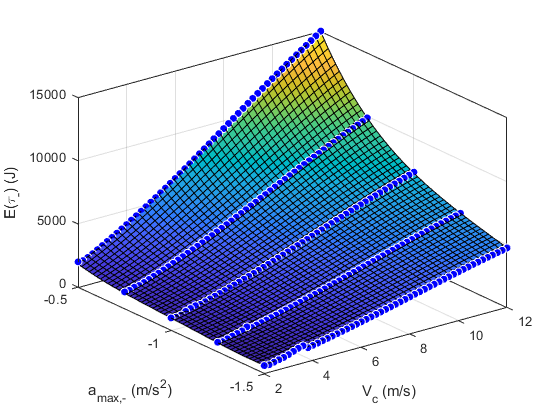}
    \end{tabular}
    \caption{Total energy use over accelerated (left) and decelerated (right) profiles as a function of $a_{max,k}$ and  $V_c$.}
    \label{fig: E_acc and E_dec}
\end{figure}

Energy consumption over accelerated flight is a function of both acceleration and cruise airspeed, requiring 3D surface fits. 
Energy over accelerated flight segment $\mathcal{T}_k$ is computed by integrating power use over the cubic spline velocity profile with maximum acceleration ($a_{max,k}$). 
Fig.\ref{fig: E_acc and E_dec} shows total energy consumed ($z$ axis) over the $Quad+Hybrid$ accelerated segment from hover to $V_c$ ($x$ axis), subject to $a_{max,k}$ ($y$ axis).
The QuadPlane accelerates in $Quad$ mode from hover to $V_{QH} = 2m/s$. It then accelerates in $Hybrid$ mode per the cubic spline to $V_c$.

Energy consumption in Fig \ref{fig: E_acc and E_dec} is a cubic function of $V_c$ and a quartic function of $|a_{max,k}|$. Consistent with Axiom 3, $E(\mathcal{T}_k)$  monotonically increases as $V_c$ increases and monotonically decreases as $|a_{max,k}|$ increases. Curve fit parameters are listed in the Appendix. 
Energy consumption is shown for all $V_c$ within $V_{QH} \leq V_c \leq V_{HP}$, i.e. $2m/s \leq V_c \leq 12m/s$ and for $|a_{max,k}| = \{0.5, 0.75, 1, 1.25, 1.5\} m/s^2$. The minimum required segment length $l_{i,i+1}$ is $4m$ when $V_c = V_{QH} = 2m/s$ and $|a_{max,k}| = 1.5 m/s^2$ per Eqs. (\ref{eq: length for a standard stop-n-go segment}) and (\ref{eq: maximum possible velocity over a segment of length l.}). The maximum required segment length $l_{i,i+1}$ is $432m$ when $|a_{max,k}| = 0.5m/s^2$ to achieve $V_c^* = V_{HP} = 12m/s$.

\subsection{Energy Optimal Traversal for the QuadPlane in No Wind}\label{Section: Energy optimal travesal for the QP}

QuadPlane curve fit equations can be applied to compute total traversal energy for given ($V_c, a_{max,+}, a_{max,-}$) values. 
Total traversal energy was computed over a range of segment lengths $10m \leq l_{i,i+1} \leq 450m$. 
Cruise airspeeds in $V_{QH} \leq V_c \leq V_{HP}$ are considered. The QuadPlane accelerates from hover to $V_c$, executes a cruise segment when $V_c$ can be reached, then decelerates tohover. Flight mode over this profile is dictated by the $(V_{QP},V_{HP})$ value pair.

\begin{figure}[!ht]
\centering
    \begin{tabular}{c c c}    
        \includegraphics[width=0.46\linewidth]{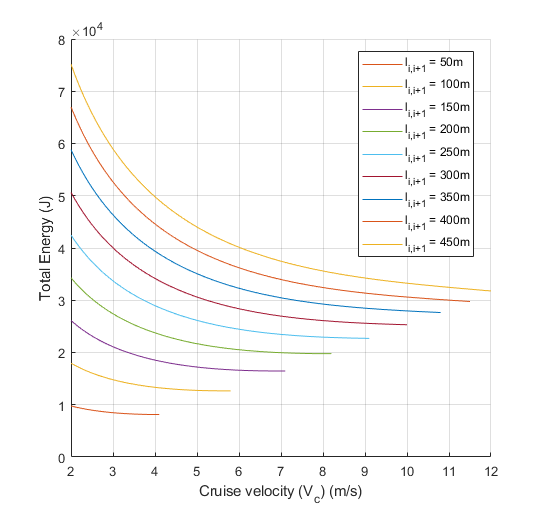} & &
        \includegraphics[width=0.46\linewidth]{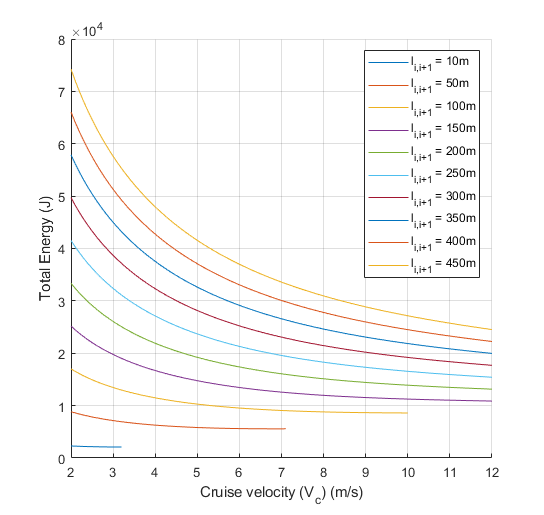} \\
        (a) $a_{max,+} = 0.5m/s^2$; $a_{max,-} = -0.5m/s^2$ & &
        (b) $a_{max,+} = 1.5m/s^2$; $a_{max,-} = -1.5m/s^2$
    \end{tabular}
    \caption{Total traversal energy as a function of $V_c$ and $l_{i,i+1}$ for given $a_{max,k}$ for feasible airspeeds.}
    \label{fig: Total energy for traversal as a function of segment length for given a_{max,k}}
\end{figure}

\begin{figure}[ht!]
    \centering
    \begin{tabular}{c c}
        \includegraphics[width = 0.47\linewidth]{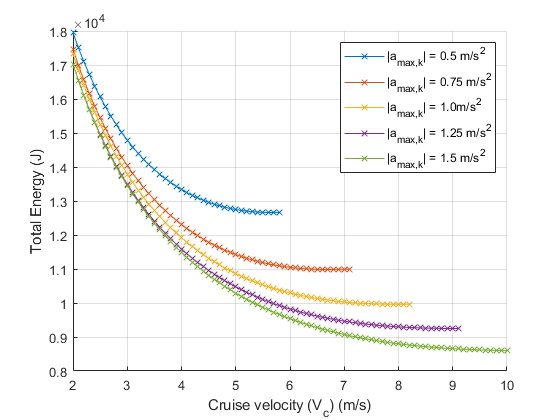}
        &  \includegraphics[width = 0.47\linewidth]{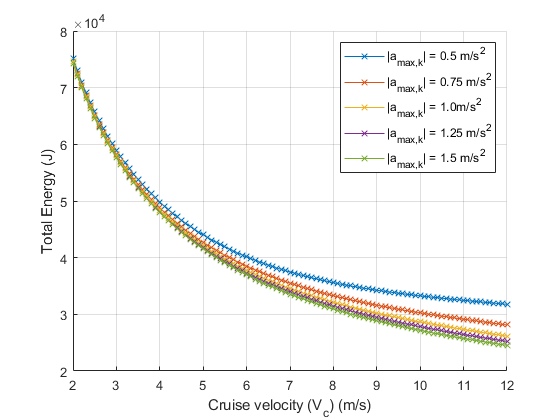}\\
        (a) $l_{i,i+1} = 100m$; & 
        (b) $l_{i,i+1} = 450m$
    \end{tabular}
    \caption{Total traversal energy as a function of $V_c$ and $a_{max,k}$ for given $l_{i,i+1}$.}
    \label{fig: total traversal energy as a function of acceleration values.}
\end{figure}

Fig. \ref{fig: Total energy for traversal as a function of segment length for given a_{max,k}} shows total energy consumed for traversals $\mathbb{T}_{i,i+1}$ of different lengths as a function of $V_c$ for two $|a_{max,k}|$ values. 
Each curve is truncated in $y$ once the maximum attainable airspeed for that segment length is reached. 
For example, when $l_{i,i+1} = 10m$ and $|a_{max,k}| = 0.5m/s^2$, the maximum achievable airspeed $V_{max} = 1.83m/s$ per Eq. \ref{eq: maximum possible velocity over a segment of length l.}, less than $V_{QH} = 2m/s$ thus not shown in the left plot.
As depicted, total energy consumption decreases as cruise airspeed increases until a minima is reached at the solution critical airspeed $V_c^{cr}$ minimizing Eq. \ref{eq: minimization problem with constraints} and optimal traversal is given by $\mathbb{T}_{i,i+1}^* = \{\mathcal{T}_+^{cr},V_c^{cr}, \mathcal{T}_-^{cr}\}$. Otherwise, optimal traversal accelerates to the maximum achievable airspeed for that segment length given by $\mathbb{T}_{i,i+1}^* = \{\mathcal{T}_+^*, \mathcal{T}_-^*\}$. 
Fig. \ref{fig: total traversal energy as a function of acceleration values.} shows total energy consumption as a function of airspeed and acceleration for two segment lengths to highlight the impact of acceleration on achievable cruise velocity. As $|a_{max,k}|$ over a given segment length increases, total energy consumption decreases and maximum achievable airspeed increases.

\begin{figure}[!ht]
    \centering
    \begin{tabular}{cc}  
        \includegraphics[width = 0.47\linewidth]{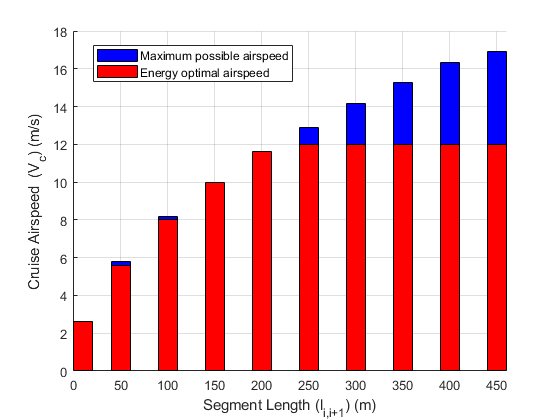} & 
        \includegraphics[width = 0.47\linewidth]{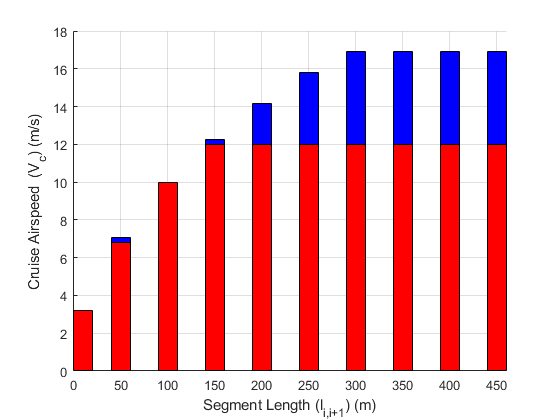} \\ 
        (a) $a_{max,+} = 1.0m/s^2$; $a_{max,-} = -1.0m/s^2$ & 
        (b) $a_{max,+} = 1.5m/s^2$; $a_{max,-} = -1.5m/s^2$ \\
    \end{tabular}    
    \caption{Optimal QuadPlane traversal airspeed as a function of segment length for two acceleration pairs.}
    \label{fig: Optimal traversal airspeed.}
\end{figure}

Fig. \ref{fig: Optimal traversal airspeed.} shows the maximum achievable airspeed as a function of traversal segment length subject to given acceleration constraints (in blue). Optimal energy traversal airspeeds are overlaid (in red). 
As illustrated, the QuadPlane cannot achieve best cruise speed $V_c^* = 12m/s$ with short segment lengths, e.g., $l_{i,i+1} < 250m$ in Fig. \ref{fig: Optimal traversal airspeed.} (left). 
In some of these cases, solution critical points exist at cruise speeds $V^{cr}_c<V_c^*$ due to the trade off between increased energy cost to accelerate to a higher airspeed versus reduced energy consumption over cruise at the higher airspeed per Eq. (\ref{eq: minimization problem with constraints}).
Optimal traversal is achieved at the boundary critical point when no solution critical points exist within constraints, per Lemma \ref{Lemma: 1}. At a boundary critical point, the QuadPlane accelerates to the maximum possible airspeed $V_{max}<V_c^{cr}$ as the segment is not long enough to reach energy minimizing cruise speed $V_c^{cr}$, followed immediately by deceleration back to hover at the destination waypoint. This case is observed for $l_{i,i+1} = 10m, 150m$ and $200m$ with $|a_{max,k}| = 1m/s^2$. When a solution critical point exists, energy cost for acceleration is offset by energy savings in cruise and thus optimal traversal includes cruise at $V_c^{cr}<V_{max}$, as seen for $l_{i,i+1} = 50m$ and $100m$ with $|a_{max,k}| = 1m/s^2$.
For long segment lengths where $V_{max}\geq V_c^*$, e.g. $l_{i,i+1}\geq 250m$ in Fig. \ref{fig: Optimal traversal airspeed.} (left), the optimal solution adopts QuadPlane's best cruise airspeed $V_c^*$ even though the QuadPlane could accelerate to its maximum thrust-limited speed $V_{lim} = 16.9 m/s$.

\subsection{QuadPlane Traversal in Steady Wind}

\begin{figure}[!ht]
    \centering
    \includegraphics[width=\linewidth]{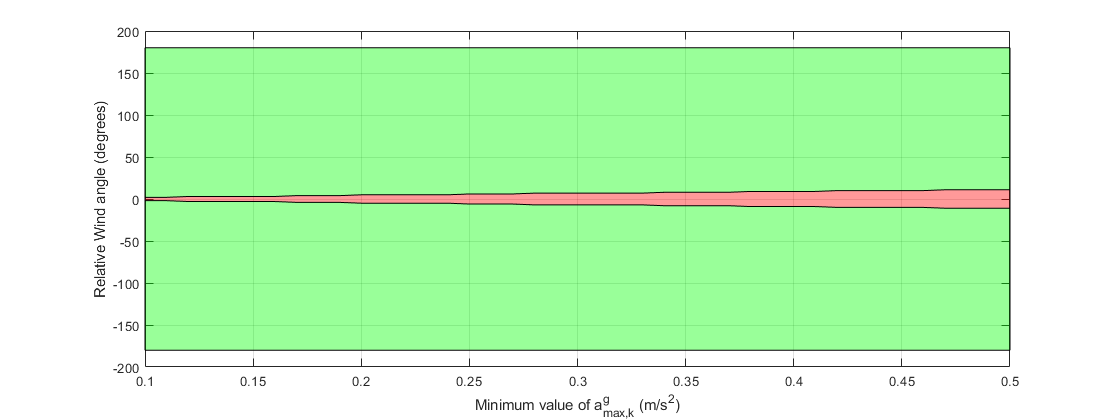}
    \caption{Straight-line traversal feasibility for steady wind magnitude of $4m/s$ as a function of relative wind angle subject to QuadPlane acceleration and heading rate constraints. Red domain indicates the need for Maneuver Primitives for Traversal.}
    \label{fig: range for STF}
\end{figure}

To investigate QuadPlane traversal in steady wind,  Algorithm \ref{alg: Straight-line traversal feasibility and computation} was first applied to identify the set of relative wind direction values where straight-line traversal is possible.  A pair of steady wind traversal cases is then presented, one where straight-line traversal is possible and another where a maneuver primitive pair is required. 
Suppose the QuadPlane transits between hover waypoints $\Vec{w}_i = [0; 0; -15]m$ and $\Vec{w}_{i+i} = [0; 500; -15]m$ in steady wind of magnitude $V^w = 4m/s$, defining a straight-line course angle $\chi = 90^\circ$. For the QuadPlane, $a^a_{lim,+} = 2m/s^2$; $a^a_{lim,-} = -2m/s^2$; $\Dot{\sigma}_{lim} = 35^\circ/s$; $\prescript{}{min}{V^g_c} = 0.01$, $|a^g_{max,k}| = 2.5m/s^2$ and $V^a_c = {V^a_c}^* = 12m/s$. Algorithm \ref{alg: Straight-line traversal feasibility and computation} was applied to assess Straight-line Traversal Feasibility $STF$ over all possible relative wind angles $\Delta \sigma$ for $\prescript{}{min}{a^g_{max,k}}$ values from $0.1m/s^2$ to $0.5m/s^2$. Fig. \ref{fig: range for STF} shows the domain where straight-line traversal is feasible in green; red shading indicates maneuver primitives are required. The range of relative wind angles requiring maneuver primitives increases with the minimum value of $a^g_{max,k}$ as expected because an increase in $\prescript{}{min}{a^g_{max,k}}$ causes a decrease in acceleration time period which in turn increases maximum heading rate until $\Dot{\sigma}_{lim}$ is reached. 
Note that in a pure tail-wind straight-line traversal is never possible and that traversal in headwind is always possible as long as the minimum ground speed constraint $\prescript{}{min}{V^g_c}$ is met. 

\begin{figure}[!ht]
    \centering
    \includegraphics[width=\linewidth]{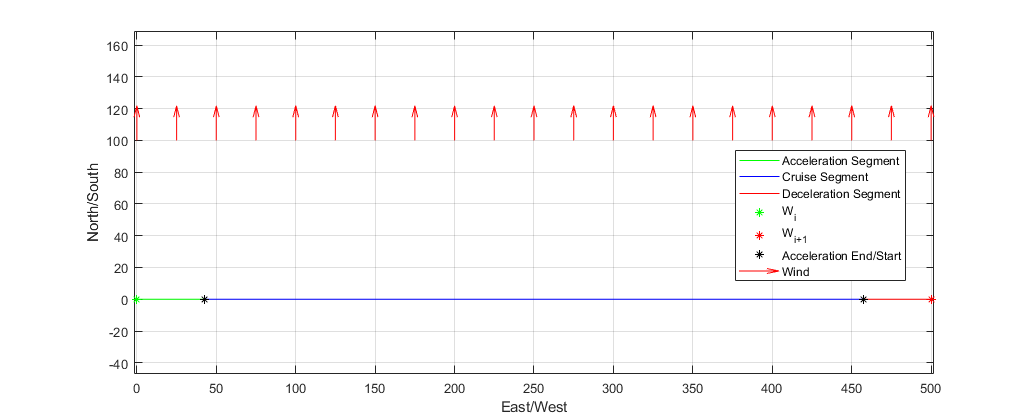}
    \caption{Straight-line traversal in a steady $4m/s$ crosswind.}
    \label{fig: st-line traversal trajectory}
\end{figure}

\begin{figure}[!ht]
    \centering
    \begin{tabular}{cc}
        \includegraphics[width=0.47\linewidth]{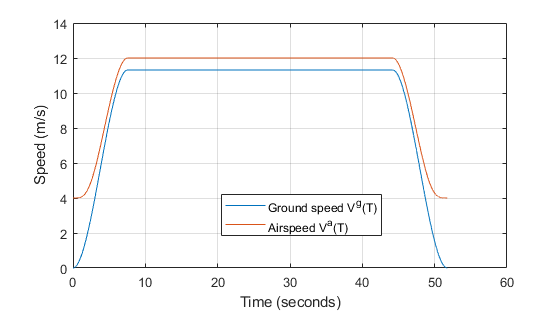}
        & 
        \includegraphics[width=0.47\linewidth]{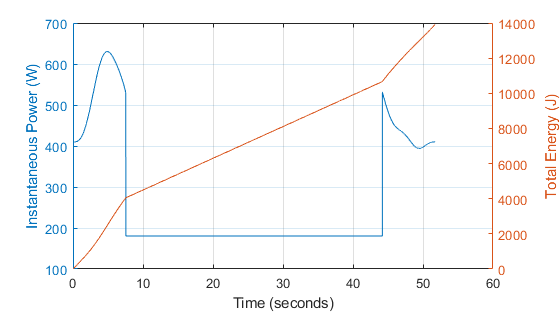}
        \\
        \includegraphics[width=0.47\linewidth]{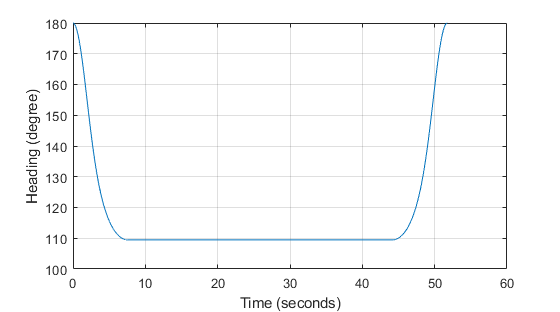} & 
        \includegraphics[width=0.47\linewidth]{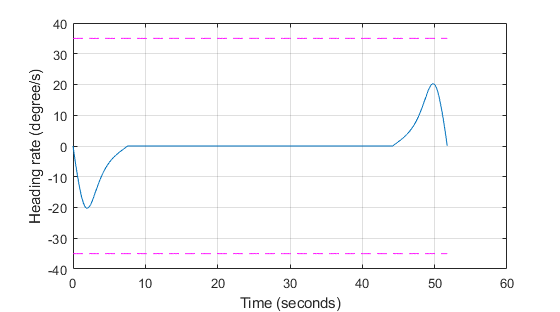}
    \end{tabular}
    \caption{Key QuadPlane parameter time histories for straight-line traversal in a steady $4m/s$ crosswind.}
    \label{fig: st-line traversal with the QP.}
\end{figure}

The first use case assumes a crosswind wind angle $\Delta \sigma = 90^\circ$, i.e. $\sigma^w = 0^\circ$. Algorithm \ref{alg: Straight-line traversal feasibility and computation} confirms straight-line traversal is feasible when $\prescript{}{min}{a^g_{max,k}}$ is $ 0.25m/s^2$. The straight-line trajectory is shown with mode and transition labels in Fig. \ref{fig: st-line traversal trajectory}. 
Time histories of ground speed $V^g(T)$ and airspeed $V^a(T)$ are shown in the top-left plot in Fig. \ref{fig: st-line traversal with the QP.}. Instantaneous power, total energy consumed, heading and heading rate time histories are also shown. 
In this case, the maximum heading rate is $\Dot{\sigma}_{max} = 20.25^\circ/s$, satisfying $\Dot{\sigma}_{max} \leq \Dot{\sigma}_{lim}$. Instantaneous power peaks during acceleration with $a^a_{max,+}$ and $a^g_{max,k}$ then drops to its minimum over the cruise segment. For cruise, the QuadPlane crabs into the wind at $19.5^\circ$ to maintain the desired straight-line ground course. Step changes in instantaneous power occur at mode switch to and from $Plane$ mode; the four vertical propulsion units are switched off after accelerating to cruise and switched back on when initiating the deceleration to hover.


\begin{table}[h!]
    \centering
    \caption{Peak Power and Energy benchmark comparisons for the crosswind straight-line traversal case study.}
    \label{tab: benchmark  st-line traversal in wind.}
    \begin{tabular}{c c c c c c c}
        Flight mode & Hover & $V^a_c$ & Crab Angle & Peak Power & Total Energy & Energy Savings\\
        \hline
        $Quad$ only & Yes & $6m/s$ & $41.8^\circ$ & $429.3 W$ & $48.50kJ$ & - \\
        $Plane$ only & No & $12 m/s$ & $19.5^\circ$ & $180.5 W$ & $7.98 kJ$ & $83.5\%$\\
        $Quad + Hybrid$ & Yes & $12m/s$ & $19.5^\circ$ & $630.4 W$ & $26.74 kJ$ & $44.9\%$\\
        
        $Quad + Hybrid + Plane$ & Yes & $12m/s$ & $19.5^\circ$ & $630.4 W$ & $13.91 kJ$ & $71.3\%$\\
    \end{tabular}
    \end{table}

To provide a benchmark comparison, total energy, peak power, and crab angle are listed in Table \ref{tab: benchmark  st-line traversal in wind.} for the steady wind traversal presented above in three additional scenarios. First, the transit is analyzed in $Quad$ mode only. For $Quad$ mode cruise airspeed $V^a_c = 6m/s$ to balance energy efficiency and stability. Second, the transit is analyzed in $Plane$ mode only with ideal cruise conditions matched at the initial and final waypoint to minimize complexity. The third additional scenario emulates the optimal multi-mode traversal solution except transition to $Plane$ mode does not occur so cruise occurs in $Hybrid$ mode at $V_c^a = 12m/s$ per Fig. \ref{fig: EPD curve fits}. As expected, the least total energy is consumed in $Plane$ mode, but this result is of limited use since it cannot be directly applied with hover waypoints.
For the three cases supporting hover waypoints, acceleration and deceleration phases are completed fastest in $Quad$ mode, but total traversal time is longest due to the reduced cruise airspeed. The proposed $Quad+Hybrid+Plane$ traversal solution provides a compromise between energy use (closest to $Plane$ only) but with hover waypoint capability.

\begin{figure}[!ht]
    \centering
    \includegraphics[width=\linewidth]{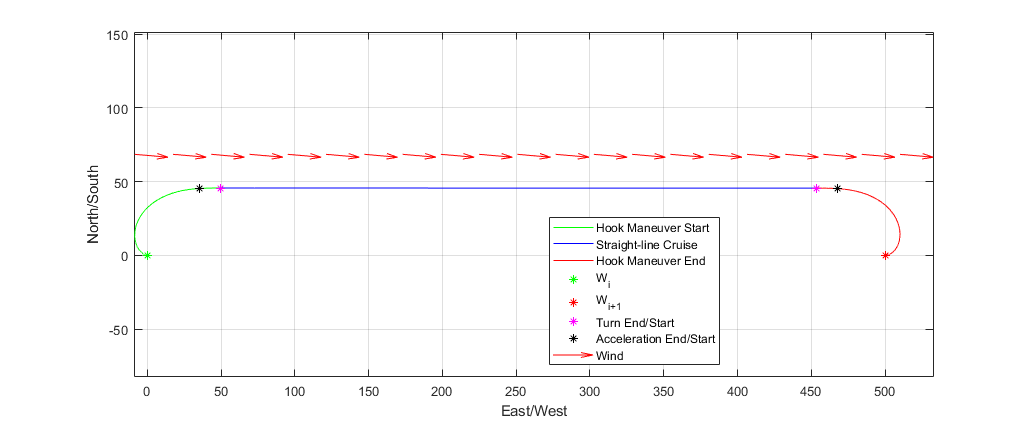}
    \caption{Cubic Spline Maneuver Primitive pair for traversal in a $4m/s$ tailwind.}
    \label{fig: trajectory for manueuver primitive traversal with the QP.}
\end{figure}

\begin{figure}[!ht]
    \centering
    \begin{tabular}{cc}
        \includegraphics[width=0.47\linewidth]{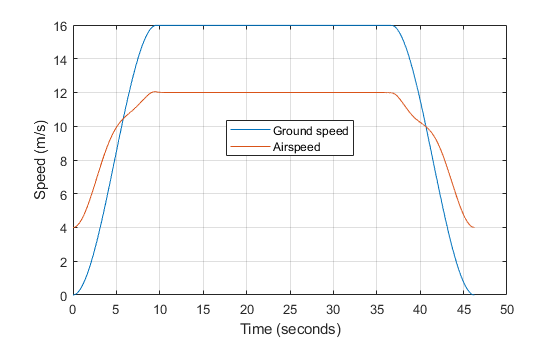}
        & \includegraphics[width=0.47\linewidth]{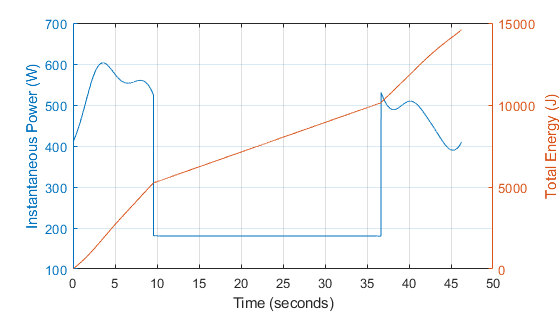}\\
        \includegraphics[width=0.47\linewidth]{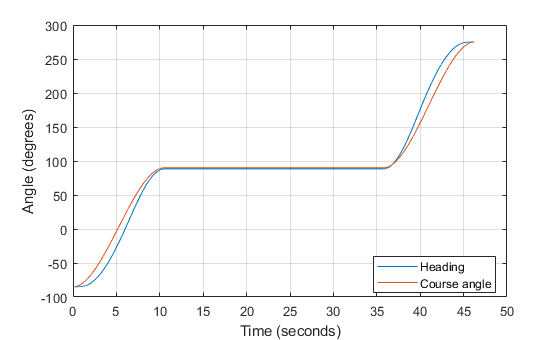} 
        & \includegraphics[width=0.47\linewidth]{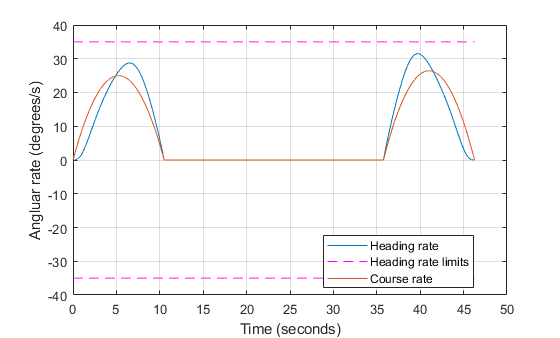}
    \end{tabular}
    \caption{Maneuver primitive traversal in steady $4m/s$ tailwind for $\Delta \sigma = -5^\circ$.}
    \label{fig: manueuver primitive traversal with the QP.}
\end{figure}

Next, consider a traversal in steady $V^w = 4m/s$ tailwind with $\sigma^w = 95^\circ$, $\Delta \sigma = -5^\circ$, and  $\prescript{}{min}{a^g_{max,k}} = 0.25m/s^2$. In this case, Algorithm \ref{alg: Straight-line traversal feasibility and computation} indicates straight-line traversal is not possible. The cubic spline maneuver primitive pair described in subsection \ref{subsection: CS maneuver primitive} produces the trajectory shown in Fig. \ref{fig: trajectory for manueuver primitive traversal with the QP.}. The acceleration maneuver starts with QuadPlane heading into the wind and course angle changes meeting the heading rate constraint. Acceleration from hover to cruise ground speed $V^g_c$ is achieved before reaching the cruise course angle, indicated by the figure's black star. Cruise course angle slightly differs from the straight-line course angle due to wind direction asymmetry. 
The deceleration cubic spline maneuver starts with a change in course angle and delays deceleration from $V^g_c$ to hover to maximize time at $V^g_c$. When the deceleration maneuver concludes, the QuadPlane points into the wind at the destination hover waypoint. 
Time histories for ground speed $V^g$, airspeed $V^a$, instantaneous power, total energy consumed, heading $\sigma$, course angle $\chi$, heading rate $\Dot{\sigma}$ and course rate $\Dot{\chi}$ are shown in Fig. \ref{fig: manueuver primitive traversal with the QP.}. Due to the tailwind, cruise ground speed exceeds cruise airspeed. Maneuver primitive pair construction ensures the heading rate constraint is met as confirmed from the heading rate plot. 

The minimum ground speed $\prescript{}{min}{V^g_c}$ constraint effectively constrains wind magnitude for successful traversal. With a headwind, cruise airspeed must be sufficiently greater than wind speed to support flight toward the destination waypoint and meet any user-defined traversal time limit. If the headwind is too strong, traversal may not be possible.

\section{Conclusion}
\label{Section: Conclusion}

This paper has optimized Lift+Cruise aircraft energy over flight mode switching and acceleration profiles for transit between hover waypoints.
Explicit constraints were derived for straight-line traversal in steady wind, and spline maneuver primitives were proposed when straight-line traversal in wind was not possible. The presented methods were applied to an experimentally validated QuadPlane eVTOL model to demonstrate optimal traversal without and with steady wind. 
In the absence of wind, maximizing acceleration and deceleration is always optimal for traversal, an expected result since energy per distance traveled decreases as airspeed increases to the energy-minimum best cruise airspeed ${V^a_c}^*$. There exists a minimum traversal distance $l_{min}^{V_c}$ for each $a_{max,k}$ such that it is not possible to exceed $V_c$. In this case, the optimal profile is to accelerate and then immediately decelerate back to hover without a cruise segment.  
With steady wind, there exists a band of relative wind angles centered around $\Delta \sigma = 0^\circ$ over which straight-line traversal between hover waypoints in steady wind is not feasible given the heading rate constraint; this band grows and shrinks with acceleration values but always exists.
Future work is needed to incorporate wind gusts and extend results to 3-D waypoint traversals and 3-D steady wind fields. This work presented an example cubic spline maneuver primitive, but optimal maneuver primitives can be derived in the future.

\section*{Appendix}

This appendix presents curve fit equations and corresponding parameters for the QuadPlane based on wind tunnel data from \cite{QP_JOA_engineering_note, QP_arXiv_journal_paper}. $RMSE$
and $R^2$ values are also tabulated. 
First, cruise power consumption is modeled with cubic and quartic functions of airspeed $V^a_c$ based on the Flight Mode $FM$: 

    \begin{equation}
        P(V^a_c) = \begin{cases}
            p_0 + p_1 V^a_c + p_2 {V^a_c}^2 + p_3 {V^a_c}^3 + p_4{V^a_c}^4; & \forall FM \in \{Q,H\} \\
            p_0 + p_1 V^a_c + p_2 {V^a_c}^2 + p_3 {V^a_c}^3; & \forall FM \in \{P\}
        \end{cases}
        \label{eq: curve fits on power vs cruise airspeed all modes}
    \end{equation}

    \begin{table}[!ht]
        \centering
        \caption{Curve fit parameters for Power Consumption as a function of Cruise Airspeed}
        \begin{tabular}{c c c c c c c c} 
            Flight Mode ($FM$) & $p_0$ & $p_1$ & $p_2$ & $p_3$ & $p_4$ & $RMSE$ & $R^2$ \\
            \hline
            Q & 270.2 & 21.66 & -18.97 & 5.822 & -0.4229 & 12.02 & 0.9998 \\
            H & 316.2 & -12.86 & 15.33 & -1.835 & 0.06427 & 10.91 & 0.9995\\
            P & -1759 & 468.8 & -42.05 & 1.246 & - & 6.75 & 0.9999        
        \end{tabular}
        \label{tab: Power curve fit params all modes}
    \end{table}

\begin{table}[!ht]
    \centering
    \caption{Surface fit parameters for Power consumption as a function of Airspeed and Instantaneous Acceleration}
    \begin{tabular}{c c c c c}
        \multirow{2}{6em}{Fit Parameter} & \multicolumn{2}{c}{Q} & \multicolumn{2}{c}{H}  \\
        \cline{2-5}
         & $a^a_+$ & $a^a_-$ & $a^a_+$ & $a^a_-$\\
        \hline
        \hline
        $p_{00}$ & 2.69 $\times10^{2}$ & 2.69$\times10^{2}$ & 2.61 $\times 10^2$ & 2.77 $\times10^{2}$\\
        $p_{10}$ & 2.92 $\times10^{1}$ & 2.94 $\times10^{1}$ & 3.90 $\times 10^1$ & 2.22$\times10^{1}$\\
        $p_{01}$ & 1.49 $\times10^{-2}$ & -7.44$\times10^{-2}$ & 3.17 $\times 10^1$ & 2.12$\times10^{1}$\\
        $p_{20}$ & -2.67 $\times10^{1}$ & -2.67$\times10^{-1}$ & -2.20 $\times10^{0}$ & 4.08$\times10^{0}$\\
        $p_{11}$ & -5.86 $\times10^{-1}$ & -2.78$\times10^{-1}$ & 5.80 $\times 10^{-1}$ & 9.63$\times10^{0}$\\
        $p_{02}$ & 1.97 $\times10^{0}$ & 1.98 $\times10^{0}$ & 1.38 $\times10^{0}$ & 8.70$\times10^{0}$\\
        $p_{30}$ & 8.76 $\times10^{0}$ & 8.80$\times10^{0}$ & 8.59 $\times 10^{-1}$ & -1.99$\times10^{-1}$\\
        $p_{21}$ & 5.78 $\times10^{-1}$& 3.75$\times10^{-1}$ & 7.90 $\times 10^{-1}$ & -2.05$\times10^{0}$\\
        $p_{12}$ & -1.10 $\times10^{-2}$ & -3.21$\times10^{-2}$ & -1.06 $\times 10^{-1}$ & -5.53$\times10^{0}$\\
        $p_{03}$ & -8.22 $\times10^{-3}$& 1.22$\times10^{-2}$ & 1.50 $\times 10^{-1}$ & -4.49$\times10^{0}$\\
        $p_{40}$ & -8.95 $\times10^{-1}$ & -9.01$\times10^{-1}$ & -1.27 $\times 10^{-1}$ & -4.57$\times10^{-2}$\\
        $p_{31}$ & -4.67$\times10^{-2}$ & -1.23$\times10^{-3}$ & -9.64 $\times 10^{-2}$ & 2.29$\times10^{-1}$\\
        $p_{22}$ & -2.76 $\times10^{-2}$& -1.97$\times10^{-2}$ & 1.41 $\times 10^{-3}$ & 8.85$\times10^{-1}$\\
        $p_{13}$ & -1.48$\times10^{-3}$ & -3.68$\times10^{-3}$ & 2.04 $\times 10^{-2}$ & 9.31$\times10^{-1}$\\
        $p_{50}$ & 2.70$\times10^{-2}$ & 2.73$\times10^{-2}$ & 5.11 $\times 10^{-3}$ & 2.78 $\times10^{-3}$\\
        $p_{41}$ & 3.31$\times10^{-3}$ & 4.06 $\times 10^{-05}$ & 3.54 $\times 10^{-3}$ & -8.68$\times10^{-3}$\\
        $p_{32}$ & 2.41$\times10^{-3}$ & 1.65$\times10^{-3}$ & -7.89 $\times10^{-5}$ & -4.03$\times10^{-2}$\\ 
        $p_{23}$ & 2.70$\times10^{-4}$ & 1.52$\times10^{-4}$ & -5.03 $\times10^{-4}$ & -4.90$\times10^{-2}$\\
        \hline
        $RMSE$ & 0.85 & 0.82 & 1.20 & 1.29\\
        $R^2$ & 0.9999 & 0.9998 & 0.9998 & 0.9996\\
    \end{tabular}
    \label{tab: Instantaneous Power surface fit params}
\end{table}

Next, QuadPlane instantaneous power consumption is modeled as a function of instantaneous acceleration $a^a$ and airspeed $V^a$. A $poly53$ surface fit is given below with corresponding parameters listed in Table \ref{tab: Instantaneous Power surface fit params}. 
    \begin{equation}
    \begin{aligned}
        P(\mathcal{T}_k) = & p_{_{00}} + p_{_{10}}{V^a} + p_{_{01}}{a^a} + p_{_{20}}{V^a}^2 + p_{_{11}} {V^a} {a^a} +  p_{_{02}}{a^a}^2 + p_{_{30}}{V^a}^3 + p_{_{21}}{V^a}^2{a^a}  + p_{_{12}}{V^a}{a^a}^2 + p_{_{03}}{a^a}^3 \\ 
        & + p_{_{40}}{V^a}^4 + p_{_{31}}{V^a}^3{a^a} + p_{_{22}}{V^a}^2{a^a}^2 + p_{_{13}}{V^a}{a^a}^3 + p_{_{50}}{V^a}^5 +p_{_{41}}{V^a}^4{a^a} + p_{_{32}}{V^a}^3{a^a}^2 + p_{_{23}}{V^a}^2{a^a}^3
    \end{aligned}
    \end{equation}

Energy per distance $\mathcal{F}$ as a function of cruise airspeed $V_c$ in zero wind conditions for $Quad$, $Hybrid$ and $Plane$ modes is described below with parameters listed in Table \ref{tab: EPD curve fit parameters.}.

\begin{equation}
    \begin{aligned}
        \mathcal{F}(V_c) = \begin{cases}
            f_0 V_c^{-f_1} + f_2, & \forall FM = \{Q,H\}\\
            f_0 + f_1 V_c + f_2 V_c^2, & \forall FM = \{P\}
        \end{cases}
    \end{aligned}
    \label{eq: EPD curve fit equations}
\end{equation}

\begin{table}[!ht]
    \caption{Energy per Distance (no wind) curve fit parameters for the QuadPlane.}
    \label{tab: EPD curve fit parameters.}
    \centering
    \begin{tabular}{c c c c c c}
        Flight mode ($FM$) & $f_0$ & $f_1$ & $f_2$ & $RMSE$ & $R^2$ \\
        \hline
        $Q$ & 235.10 & 1.48 & 52.27 & 2.74 & 0.997\\
        $H$ & 277.30 & 0.53 & -27.93 & 1.37 & 0.998\\
        $P$ & 0.65 & -16.37 & 117.50 & 0.12 & 0.999
    \end{tabular}    
\end{table}

Energy consumption is analyzed in no wind over an accelerated segment following a cubic spline profile from hover to cruise airspeed $V_c$ with maximum acceleration $a_{max,k}$. A $poly 34$ surface is fit with parameters listed in Table \ref{tab: E_acc and E_dec surface fit parameters.}.

\begin{equation}
\begin{aligned}
    E(\mathcal{T}_k) = & p_{_{00}} + p_{_{10}}{V_c} + p_{_{01}}{a_{max,k}} + p_{_{20}}{V_c}^2 + p_{_{11}} {V_c} {a_{max,k}} +  p_{_{02}}{a_{max,k}}^2 + p_{_{30}}{V_c}^3 + p_{_{21}}{V_c}^2{a_{max,k}}  + p_{_{12}}{V_c}{a_{max,k}}^2 \\
    & + p_{_{03}}{a_{max,k}}^3 + p_{_{31}}{V_c}^3{a_{max,k}} + p_{_{22}}{V_c}^2{a_{max,k}}^2 + p_{_{13}}{V_c}{a_{max,k}}^3 + p_{_{04}}{a_{max,k}}^4
\end{aligned}
\end{equation}

\begin{table}[ht!]
    \caption{Surface fit parameters for Energy consumption during an accelerated segment.}
    \label{tab: E_acc and E_dec surface fit parameters.}
    \centering
    \begin{tabular}{c c c }
        Fit parameters & $\mathcal{T}_+$ & $\mathcal{T}_-$\\
        \hline
        $p_{00}$ & 3.1 $\times 10^3$ & 3.7$\times 10^3$\\
        $p_{10}$ & 2.1 $\times 10^3$ & 1.8$\times 10^3$\\
        $p_{20}$ & 167.6 & 187.4\\
        $p_{30}$ & -6.2  & -7.2\\
        $p_{01}$ & -15.8 $\times 10^3$ & 15.7$\times 10^3$\\
        $p_{02}$ & 29.4 $\times 10^3$ & 28.1$\times 10^3$\\
        $p_{03}$ & -22.4 $\times 10^3$ & 21.1$\times 10^3$\\
        $p_{04}$ & 5.9 $\times 10^3$ & 5.6$\times 10^3$\\
        $p_{11}$ & -4.7 $\times 10^3$ & 4.5$\times 10^3$\\
        $p_{12}$ & 3.8 $\times 10^3$ & 3.7$\times 10^3$\\
        $p_{21}$ & -103.8 & 119.9\\
        $p_{22}$ & 12.5 & 15.9\\
        $p_{31}$ & 3.2 & -3.7\\
        $p_{13}$ & -1.0 $\times 10^3$ & 1.0$\times 10^3$\\
        \hline
        $RMSE$ & 18.47 & 40.16\\
        $R^2$ & 0.9999 & 0.9998\\
    \end{tabular}
\end{table}

\newpage
\bibliography{references}   

\end{document}